\documentclass[twocolumn]{fairmeta}

\title{Spectral Filters, Dark Signals, and Attention Sinks}

\author[]{Nicola Cancedda}

\affiliation[]{FAIR at Meta}

\abstract{Projecting intermediate representations onto the vocabulary is an increasingly popular interpretation tool for transformer-based LLMs, also known as the \textit{logit lens} \citep{nostalgebraist2020}. We propose a quantitative extension to this approach and define \textit{spectral filters} on intermediate representations based on partitioning the singular vectors of the vocabulary embedding and unembedding matrices into bands. We find that the signals exchanged in the tail end of the spectrum are responsible for attention sinking \citep{xiao2023efficient}, of which we provide an explanation. We find that the loss of pretrained models can be kept low despite suppressing sizeable parts of the embedding spectrum in a layer-dependent way, as long as attention sinking is preserved. Finally, we discover that the representation of tokens that draw attention from many tokens have large projections on the tail end of the spectrum.}

\date{\today}
\correspondence{Nicola Cancedda at \email{ncan@meta.com}}

\usepackage{graphicx} %
\usepackage{amsmath}
\usepackage{makecell}
\usepackage{breqn}

\begin{document}

\maketitle

\section{Introduction}
\label{section:intro}

Large foundation models dominate the state of the art in numerous AI tasks. While we understand how these models work in terms of elementary operations, and black-box evaluations help characterize observable behaviours, we lack a clear understanding of the connection between the two.

There is a growing body of work providing insights into properties of model components, e.g. \citep{voita2019analyzing,pimentel2020information,voita2020information,geva2022transformer,meng2022locating,voita2023neurons}, as well as identifying and explaining fundamental phenomena, often with the support of simple models \citep{elhage2021mathematical,elhage2022toy,olsson2022context,todd2023function}.

Most recent works assign a central role to the model's \textit{residual stream} (RS) as the shared communication channel between model components. In this perspective, the probability distribution of a token is initialised from the projection of the embedding of the previous token through the unembedding matrix, and receives additive updates from attention heads and MLP components, each reading from the residual stream of the same or previous tokens. The role played by components is interpreted projecting their contribution on the probability distribution over vocabulary items, in what is referred to as the \textit{logit lens} \citep{nostalgebraist2020,geva2020transformer}. We extend this approach and introduce \textit{logit spectroscopy}, the spectral analysis of the content of the residual stream and of the parameter matrices interacting with it. Equipped with this tool, we look at the part of the residual stream spectrum that is most likely to be neglected by the logit lens: the linear subspace spanned by the right singular vectors of the unembedding matrix with the \textit{smallest} singular values. Drawing an analogy with ``dark matter'' in astrophysics, that interacts with light only indirectly, we dub projections onto this subspace \textit{dark} parameters, features, activations etc.

\begin{figure}
    \centering
    \includegraphics[width=\linewidth]{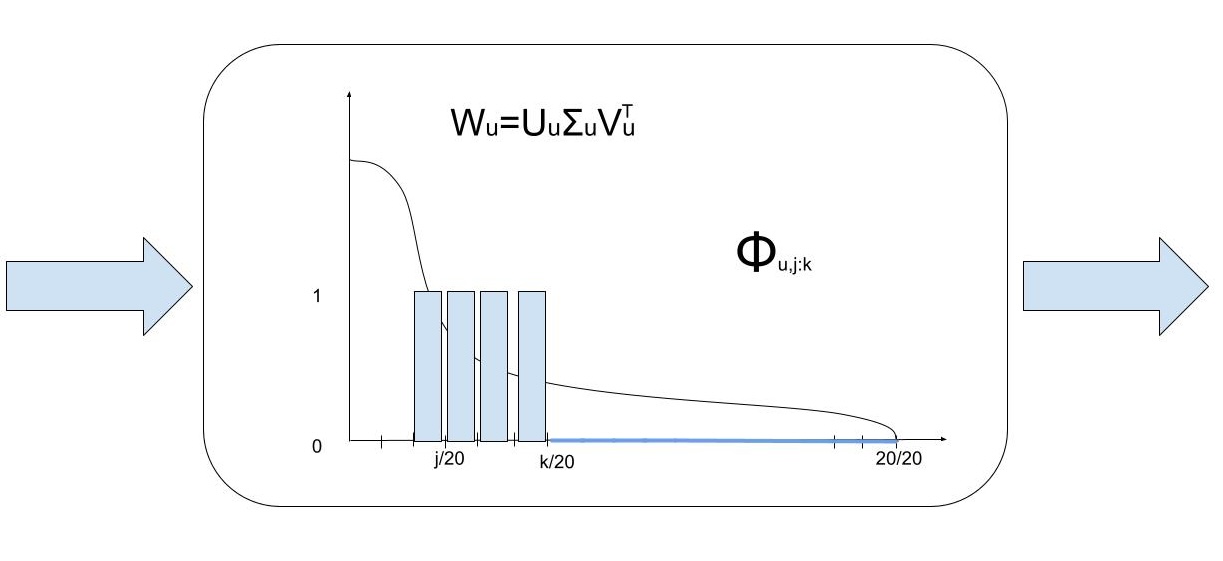}
    \caption{Spectral filters project signals exchanged between components onto selected subspaces as defined by the spectral decomposition of the vocabulary embedding and unembedding matrices of the model.}
    \label{fig:phi_filter}
\end{figure}

We were motivated by the thought that LLMs could learn to use signals in the dark linear subspace to maintain global features responsible for long-range dependencies while minimizing their interference with the next token prediction. We discovered instead that dark signals are instrumental to implementing the recently described phenomenon of \textit{attention sinks} \citep{xiao2023efficient}, of which we provide a detailed account. We also show that the negative log likelihood of pretrained models can be kept low despite suppressing large swaths of the unembedding spectrum, as long as the dark signals required for attention sinking are untouched. Finally, we find a significant positive correlation between the average attention received by a token and the relative prevalence of dark signals in its residual stream.

\section{The LLaMa2 models}
\label{sec:llama2}

We chose LLaMa2 models \citep{touvron2023llama} as the object of our study as they were the most competitive models with open-access weights at the time we started this work. In particular we studied pretrained models (without instruction fine-tuning) with 7B, 13B, and 70B parameters.

Table \ref{tab:llama2_family} presents the key dimensional hyperparameters of these models.
\begin{table*}[]
    \centering
    \begin{tabular}{|c|c|c|c|c|c|}\hline
        Model & Layers & RS dim & MLP dim & Attn heads & KV heads\\\hline
        LLaMa2 7B & 32 & 4096 & 11008 & 32 & 32 \\
        LLaMa2 13B & 40 & 5120 & 13824 & 40 & 40 \\
        LLaMa2 70B & 80 & 8192 & 14336 & 64 & 8 \\\hline
    \end{tabular}
    \caption{Dimensional hyper-parameters of the LLaMa2 models. The 70B model uses grouped-query attention, with 64 attention heads sharing 8 key and value matrices.}
    \label{tab:llama2_family}
\end{table*}
\noindent
All LLaMa2 models share the same tokenizer, with 32,000 vocabulary items. Table \ref{tab:notation}, in Appendix \ref{sec:notation}, summarizes the standard notation we use to refer to model components.

We point out a few differences between LLaMa2 models and the original transformer architecture \citep{vaswani2017attention}.
RMSnorm \citep{zhang2019root} is applied \textit{before} every attention component, MLP component, and the final unembedding projection. RMSnorm includes a learned rescaling vector that is applied after the normalization proper. In all cases, we absorb this rescaling vector into the matrices downstream from the rescaling: this is mathematically equivalent and simplifies the analysis.

LLaMa2 models use SwiGLU activation functions \citep{shazeer2020glu}:

\begin{dmath}
    \text{FFN}_{\text{SwiGLU}} (x,W_1,W_2,W_3) = (\text{Swish}_1(xW_1)\otimes xW_3)W_2
\end{dmath}

\noindent
This means that MLP components have three parameter matrices instead of the more familiar two.

Finally, LLaMa2 models use rotary positional embeddings \citep{su2023roformer}. While important for the functioning of the model, their effect is independent of the phenomena we are focusing on.

\section{Model parameter properties}
\label{sec:parameter_properties}

\begin{figure}[!t]
    \begin{centering}
        \includegraphics[width=0.49\textwidth]{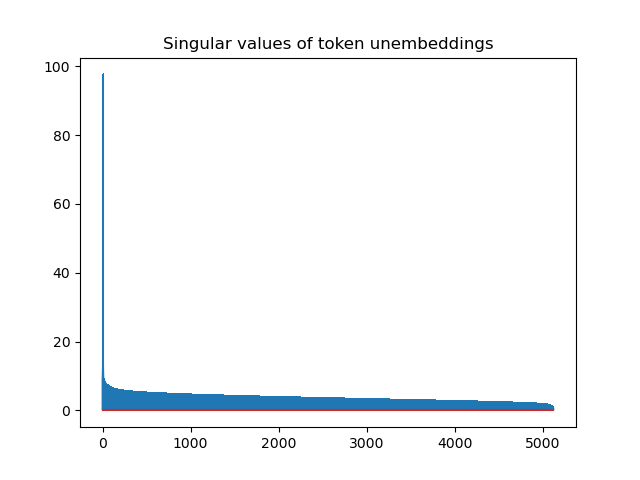}
        \caption{Distribution of the singular values of the unembedding matrix $W_u$ of LLaMa2 13B. The \textit{U-Dark} subspace is the one spanned by the last 5\% right singular vectors.}
        \label{fig:wu_sv_distribution}
    \end{centering}
\end{figure}

Let $W_u = U_u\Sigma_u V_u^{\top}$ be the singular value decomposition of the vocabulary unembedding matrix (and similarly for $W_e$). Figure \ref{fig:wu_sv_distribution} shows the distribution of the singular values (SVs) of $W_u$ for LLaMa2 13B. There is a single large SV\footnote{Projecting unembeddings on the first singular vector shows that it is highly representative of token frequency.} followed by a tail that declines only at the very end (the distribution for LLaMa2 7B and LLaMa2 70B is similar). The SV distribution of the $W_e$ embedding matrix is also similar, but the top SV is only twice as large as the second one, with a longer ``head'' of relatively large SVs.

We use the adjective \textit{U-dark} (\textit{E-dark}) to characterize anything that happens in the linear subspace spanned by the 5\% right singular vectors (RSVs) of $W_u$ ($W_e$) with the smallest SVs, the dark basis.

We gain an initial insight into whether dark signals are exchanged by projecting rows and columns of parameter matrices on the dark bases. $W_k$, $W_q$, and $W_v$ project the residual stream onto either the latent space used to compute attention scores or the latent space used to compute the attention output, so they ``read'' from the residual stream; similarly $W_1$ and $W_3$ map from the residual stream onto the activation layer of the MLP components. We project the columns of these matrices on the RSVs of $W_u$ to estimate and visualize their aptitude to read from the dark subspace\footnote{We adopt the convention that input vectors are row vectors and are multiplied by parameter matrices on the right.}. Conversely, $W_o$ and $W_2$ map from latent spaces back into the residual stream, therefore we project their rows to check their aptitude to write into the dark subspace. We computed and plotted the norms of the $d$ vectors of dimension $d_h$ or $d_m$ obtained with these projections, e.g.:

\begin{equation}
j_i =  
\begin{cases}
    ||(V_{u}^{\top})_iW_y||_2, & y\in \{k, q, v, 1, 3\}\\
    ||(V_{u}^{\top})_iW_y^{\top}||_2, & y\in \{o,2\}
\end{cases}
\end{equation}

We discovered a great variety in where, in the bases formed by the RSVs of $W_u$ and $W_e$, model components are equipped to read from and write into. 
\begin{figure}
    \centering
    \parbox{0.21\textwidth}{
        \includegraphics[width=0.24\textwidth]{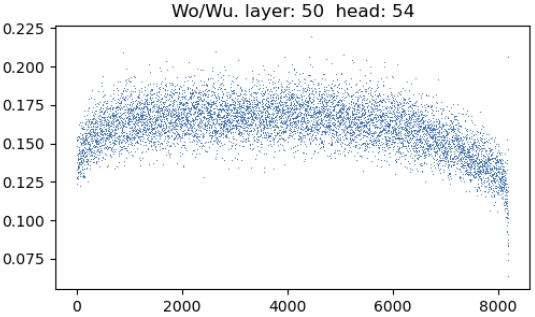}
    }
    \quad
    \parbox{0.21\textwidth}{
        \includegraphics[width=0.24\textwidth]{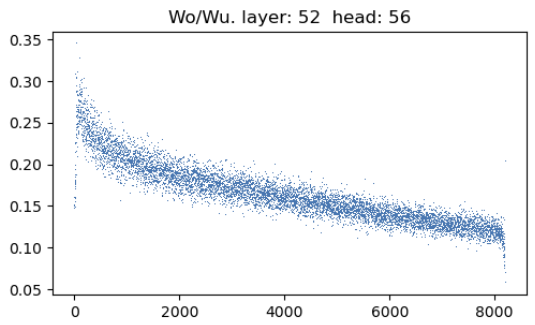}
    }
    \parbox{0.21\textwidth}{
        \includegraphics[width=0.24\textwidth]{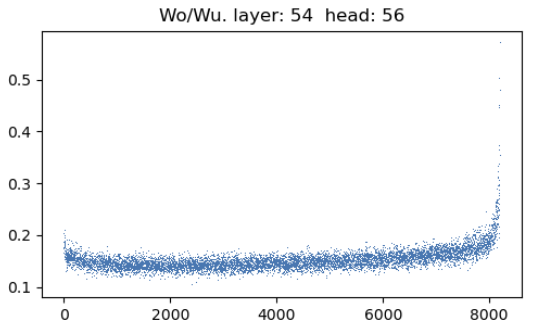}
    }
    \quad
    \parbox{0.21\textwidth}{
        \includegraphics[width=0.24\textwidth]{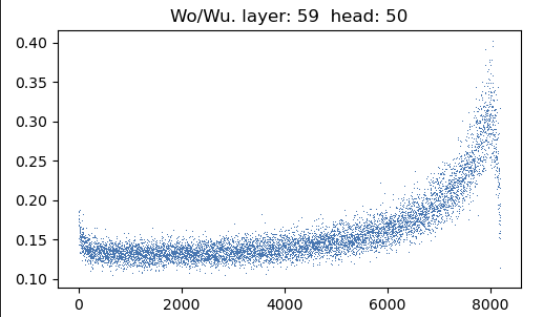}
    }
    \caption{The projections of four $W_o$ matrices of LLaMa2 70B on the RSVs of $W_u$. Different heads are equipped to write into different subspaces, with some targeting the dark subspace.}
    \label{fig:wo_proj_wu_tile}
\end{figure}
Figure \ref{fig:wo_proj_wu_tile} gives a sense of such variety.
\begin{figure}
    \centering
    \includegraphics[width=0.49\textwidth]{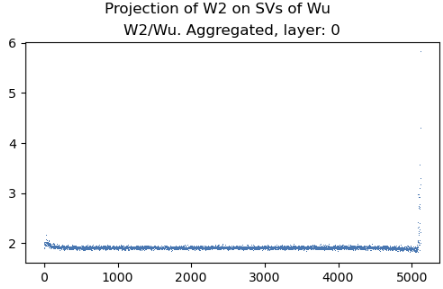}
    \caption{The projection of the rows of $W_2$ at L0 of LLaMa2 13B on the RSVs of $W_u$. Note the large values at the very right end of the spectrum, indicating the ability to write in the U-Dark space.}
    \label{fig:w2_proj_wu}
\end{figure}

The projections of MLP matrices also clearly indicate that some MLP components can write mostly into the dark subspace (see Fig.\ref{fig:w2_proj_wu} for the projection of 13B/L0/$W_2$ on $W_u$).

There are therefore components that are equipped for communicating through the dark subspace. In the following section we explore if such communication actually takes place and how it manifests itself.

\section{Spectral filtering}
\label{sec:spectral_filtering}

One possible explanation for the existence of components that communicate through the dark subspace is that they are useless, and the model learned to divert their output to subspaces with little bearing on the vocabulary logits. To see if this is the case, we perform a series of experiments, where we patch some of the intermediate representations by projecting them onto the RSVs of $W_u$ and $W_e$ with largest singular vectors, therefore removing dark signals. We measure the average negative log-likelihood (NLL) of tokens in a sample of prompts: if dark signals are irrelevant noise then we expect NLL to be largely unaffected when they are filtered out.

We split the singular vectors of $W_u$ into 20 \textit{bands}: \{$V_{u,1},\ldots,V_{u, 20}$\}, of cardinality $d/20$, and similarly for $W_e$. $V_{u,1}$ contains the 5\% of singular vectors with largest singular values, $V_{u,2}$ the next 5\%, and so on. Let $V_{u,j:k}$ be the matrix obtained concatenating $[V_{u,j}\ldots V_{u,k}]$, and (Fig. \ref{fig:phi_filter}) let $\Phi_{u,j:k} = V_{u,j:k}V_{u,j:k}^{\top}$ (similarly for $W_e$). We can then project any $d$-dimensional vector onto the U-dark (E-dark) space by multiplying them by $\Phi_{u,20:20}$ (resp. $\Phi_{e,20:20}$). 

\begin{figure}
    \centering
    \includegraphics[width=\linewidth]{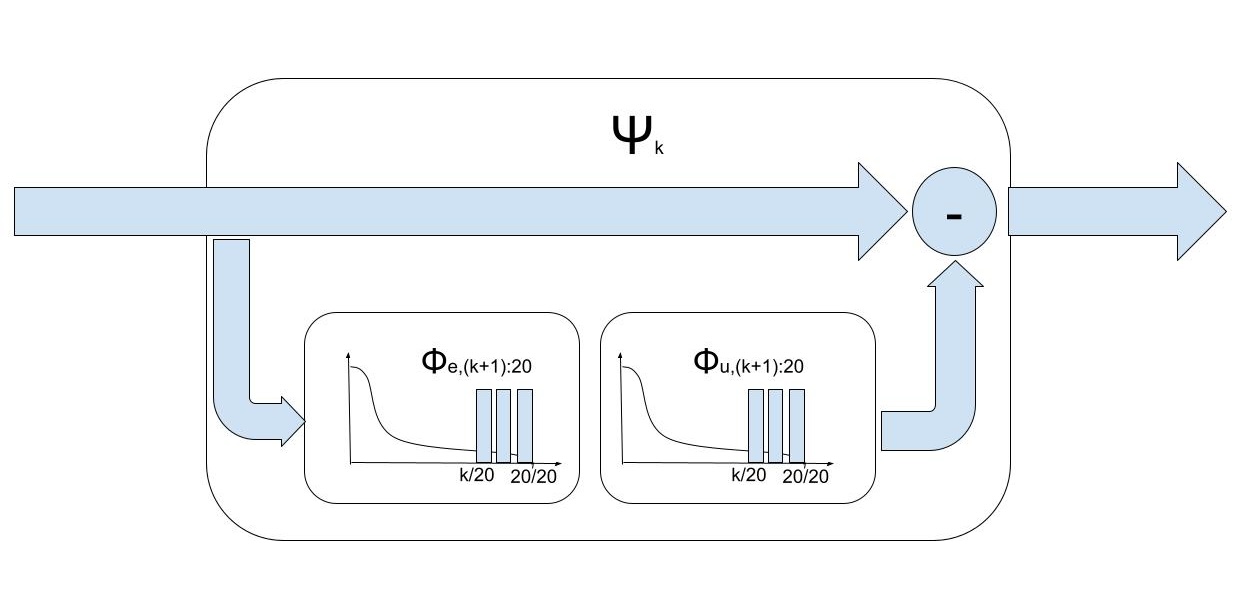}
    \caption{$\Psi$ filters project vectors onto subspaces that are dark according to both the embedding and the unembedding matrix decomposition.}
    \label{fig:psi_filter}
\end{figure}
We form a hierarchy of nested filters $\Phi_{u,1:k}$: multiplying a vector by $\Phi_{u,1:k}$ projects it onto the subspace of the $kd/20$ ``least dark'' dimensions. We also define filters that combine singular vectors of $W_u$ and $W_e$ (Fig. \ref{fig:psi_filter}):

\begin{equation}
    \Psi_k = (I - \Phi_{e,(k+1):20}\Phi_{u,(k+1):20}), k = 1, \ldots, 19
\label{eq:PsiK}
\end{equation}

\noindent
and we set $\Psi_{20} = I$. Multiplying a vector by $\Psi_k$  filters away projections onto subspaces of both $W_e$ and $W_u$ that get darker as $k$ grows. $\Psi_{19}$ filters out only vector components in the 5\% dark subspace of \textit{both} $W_u$ and $W_e$. These definitions make $\Phi_{u,k}$, $\Phi_{e,k}$, and $\Psi_k$ comparable when looking at the fraction of kept/discarded singular vectors, although, since it discards a double projection, $\Psi_k$ retains more information.
We create a dataset (ccnet-405) of 405 prompts (13,268 tokens) in English, from CCNET \citep{wenzek2019ccnet}. 
\begin{figure}[t]
    \centering
    \includegraphics[width=0.49\textwidth]{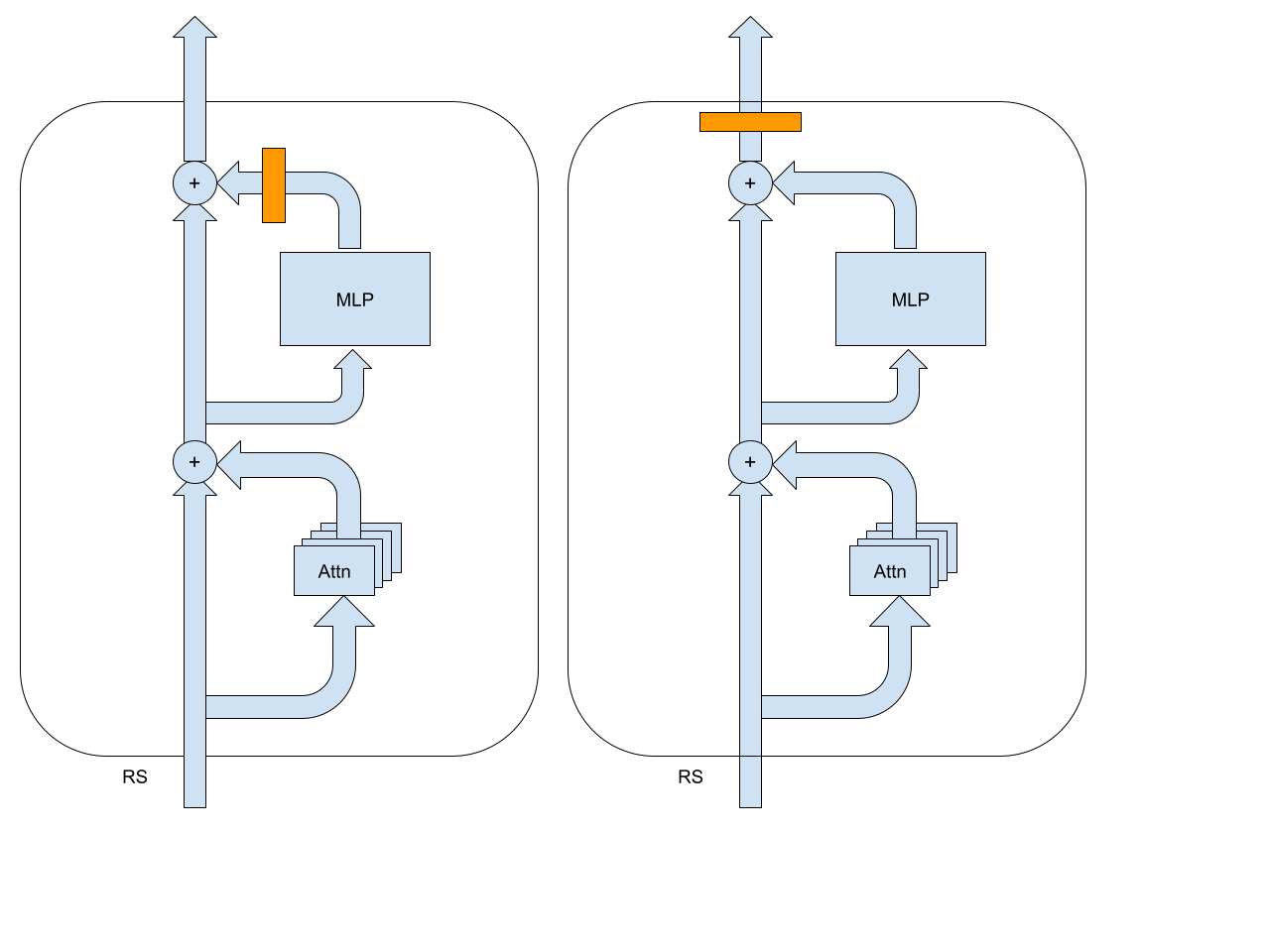}
    \caption{The two positions where we applied spectral filters: on the output of MLP layers, one at a time (left) or on the residual stream after a complete layer (right).}
    \label{fig:bottlenecks}
\end{figure}
We ran inference on ccnet-405 applying spectral filters to the output of MLP components, one by one (Fig. \ref{fig:bottlenecks} (left)) and measured the average NLL. We similarly measure NLL when filtering the RS after the contributions of a given target layer have been added to it (Fig. \ref{fig:bottlenecks} (right)). See App. \ref{sec:implementation_details} for implementation details.

\subsection{Discussion}
\label{subsec:spectral_filtering_discussion}

When filtering MLP layers of LLaMa2 13B, the effects of masking L0 and L3 dwarf all others.

\begin{figure}
    \centering
    \includegraphics[width=\linewidth]{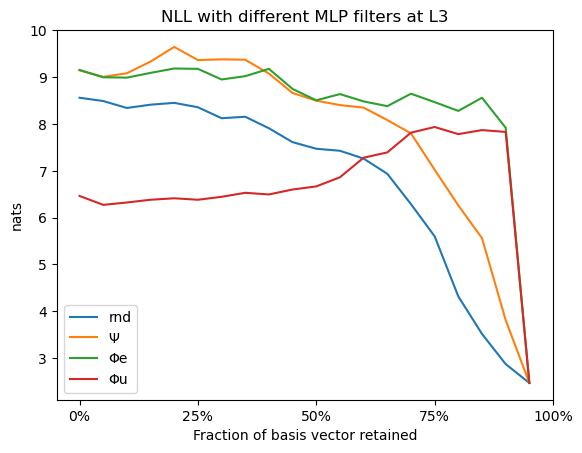}
    \caption{The effect of filtering 13B/L3/MLP with the filters defined in Section \ref{sec:spectral_filtering}. 'Rnd' indicates filtering projection on subsets of a random orthonormal basis, for reference.}
    \label{fig:mlp_filtering_L3}
\end{figure}

When filtering the output of the MLP at L3 (Fig. \ref{fig:mlp_filtering_L3}) the loss remains poor until the last 5\% of the spectrum is included for both the $\Phi$ spectral filters. The fact that the $\Psi$ curve is always above the random filter indicates that this MLP exerts strong influence operating in the dark subspace. See Appendix \ref{app:additional_spectral_filtering_plots} for a discussion of 13B/L0/MLP.

Figure \ref{fig:filter_7b_70b_mlp_one} in App. \ref{app:additional_spectral_filtering_plots} shows that MLP components with a similar propensity for writing dark signals exist also in LLaMa2 7B (L1) and 70B (L2 and L8).

\begin{figure*}
    \centering
   \includegraphics[width=\linewidth]{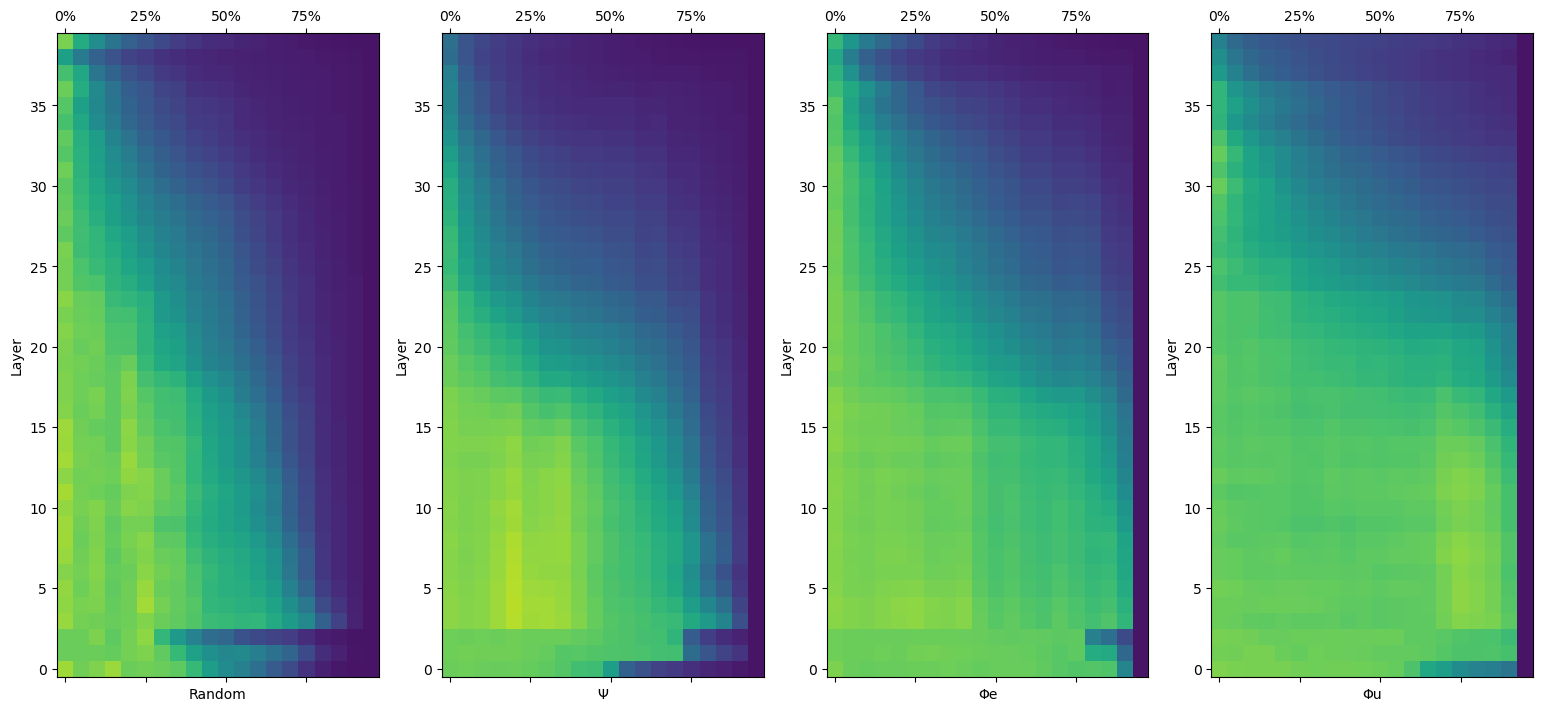}
    \caption{NLL of LLaMa2 13B on ccnet-405 when filtering the residual stream after a given layer (Y-axis), retaining an increasing number of SVs (X-axis) of (c) $W_e$ or (d) $W_u$. (b) filters from the residual stream only its double projection onto both the $W_e$ and $W_u$ dark spaces. (a) shows, for comparison, the effect of adding more and more dimension in a random orthogonal base. See Fig.\ref{fig:filter_7b_70b_rs_one} for similar heatmaps for LLaMa2 7B and 70B.}
    \label{fig:filter_13b_rs_one}
\end{figure*}

We also look at what happens to the negative log-likelihood when filtering the residual stream after a given layer (Figures \ref{fig:bottlenecks} (right) and \ref{fig:filter_13b_rs_one}).
Even a mild bottleneck after L3 results in a significant increase in NLL, in line with the observation that the MLP in L3 writes in the dark space. If we filter only signals that are dark to both $W_e$ and $W_u$ (Fig. \ref{fig:filter_13b_rs_one}(b)), we see bottlenecks continuing to be more harmful than random direction removal (Fig.\ref{fig:filter_13b_rs_one}(a)) up until L20-25.

Dark signals exist and play an important role in achieving low perplexity, but what is this role?

\section{Attention sinks}
\label{sec:attention_sinks}

\cite{xiao2023efficient} describe \textit{attention sinks}: the special Beginning of Sentence (BoS) 'Token 0' receives a disproportionate amount of attention.\footnote{\citet{xiao2023efficient} include the first four tokens in their operational definition of attention sinks, for achieving effective streaming decoding. In our observations the BoS token is by far the one attracting the most attention.} This happens because often an attention head should not be activated in a given context, but the normalisation in the attention scores forces a constant amount of attention to be distributed to previous tokens. The model therefore learns to sink excess attention by allocating it to the BoS token. Since attention heads map content from the residual stream being attended to to that of the present token, the content imported when sinking attention should interfere as little as possible with the probability distribution of the next token.

\begin{figure}
    \centering
    \includegraphics[width=\linewidth]{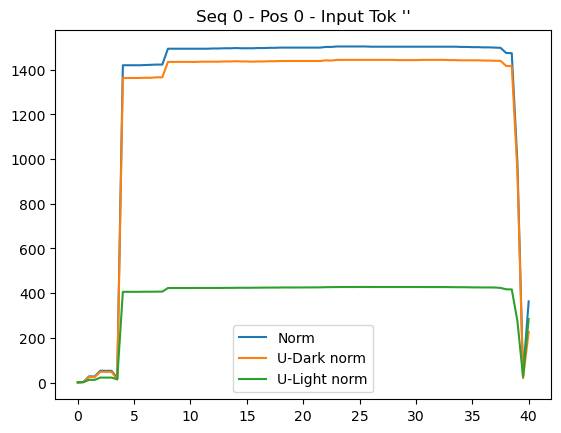}
    \includegraphics[width=0.48\linewidth]{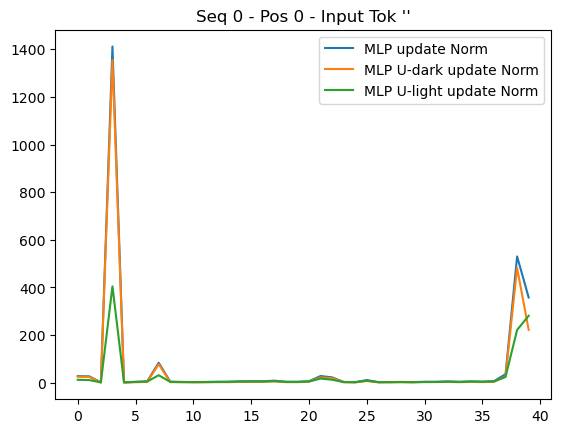}
    \includegraphics[width=0.48\linewidth]{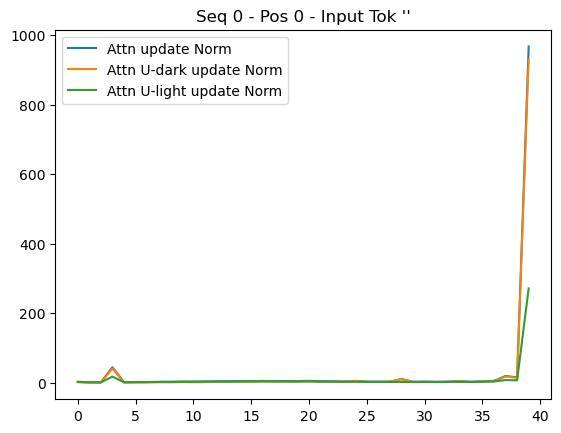}
    \caption{(Top) The composition of the RS of the BoS token for LLaMa2 13B as a function of the layer. (Bottom left) The norms of the contribution of MLP layers to the BoS RS. (Bottom right) The norms of the contribution of Multi-Head Attention components to the BoS RS.}
    \label{fig:bos}
\end{figure}

Figure \ref{fig:bos} shows how the norm of the Token 0 residual stream progresses over layers, and the  contributions from MLPs and MHAs components. We plot the overall norm, and the norms of the projection on the U-Dark space and of the projection on its orthogonal complement (\textit{U-Light}).

After an initial phase of input enrichment that apparently does not need an attention sink, the MLP at L3 blasts off a vector of large norm and almost completely U-dark. This vector acts as an attention collector for heads in need of a sink, and is kept around until the last few layers, where the combined action of MLPs and attention heads first erases it and then replaces it with the vector that encodes the probability distribution of the model over generation-initial tokens.

\begin{figure}[t]
    \centering
    \includegraphics[width=0.49\textwidth]{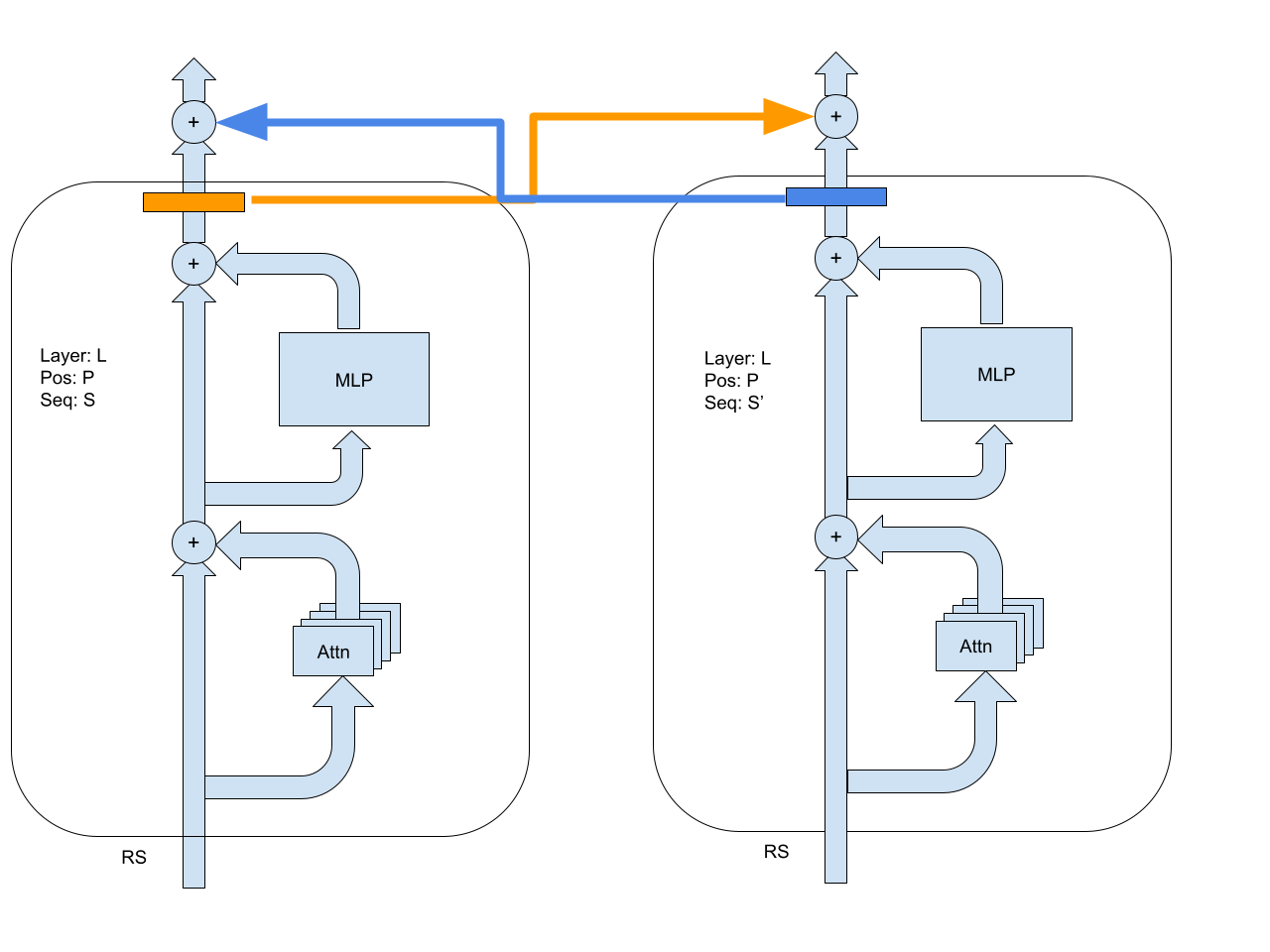}
    \caption{Shavings swap experiment: rather than being suppressed, the subspaces blocked by spectral filters are swapped with the corresponding ones from a token at the same position but in a different sample. This swap perturbs all residual streams except the one for Token 0, since this is identical for all samples due to the autoregressive nature of the model.}
    \label{fig:patch_swap}
\end{figure}

We confirmed that dark signals are primarily used to sink attention with an additional experiment (Fig. \ref{fig:patch_swap}). We apply spectral filters at the exit of a layer but, rather than suppressing the filtered component, we add it into the corresponding residual stream of a different sample, at the same token position and layer. Since the content of the RS of the BoS is the same irrespective of the input, this procedure leaves it intact, while perturbing all other RSs.
\begin{figure}
    \centering
    \includegraphics[width=\linewidth]{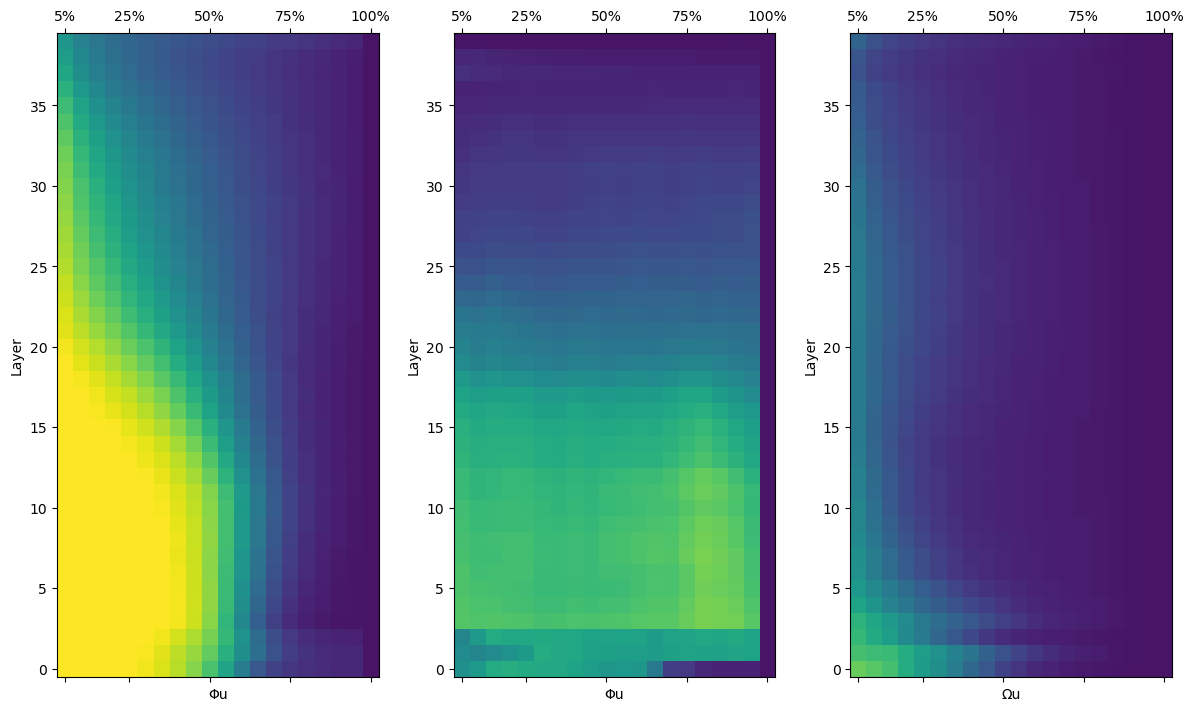}
    \caption{NLL of LLaMa2 13B on ccnet-405 when filtering the residual stream after a given layer (Y-axis). (Left:) Swapping the filtered vector components between RS at the same layer and token position, but in a different sample, perturbing all RSs except the BoS one. (Middle:) Filtering only the residual stream of the BoS token. (Right:) Applying the sink-preserving spectral filters $\Omega_{u,k}$ to a section of the residual stream of LLaMa2 13B right after a layer.}
    \label{fig:filter_13b_patch_bos_omega}
\end{figure}

The resulting NLL heatmap is in Fig.\ref{fig:filter_13b_patch_bos_omega} (left). Unlike in those in Fig. \ref{fig:filter_13b_rs_one}, there is no step-decrease in NLL when the last 5\% of singular vectors is added. Coversely, if we apply spectral filters \textit{only} to Token 0, the step-decrease is clearly visible (Fig.\ref{fig:filter_13b_patch_bos_omega} (middle)).
We conclude therefore that the primary function of the dark subspace is to enable the crucial attention sink mechanism. 

\section{Sink-preserving spectral filters}
\label{sec:omega_filters}

\begin{figure}
    \centering
    \includegraphics[width=\linewidth]{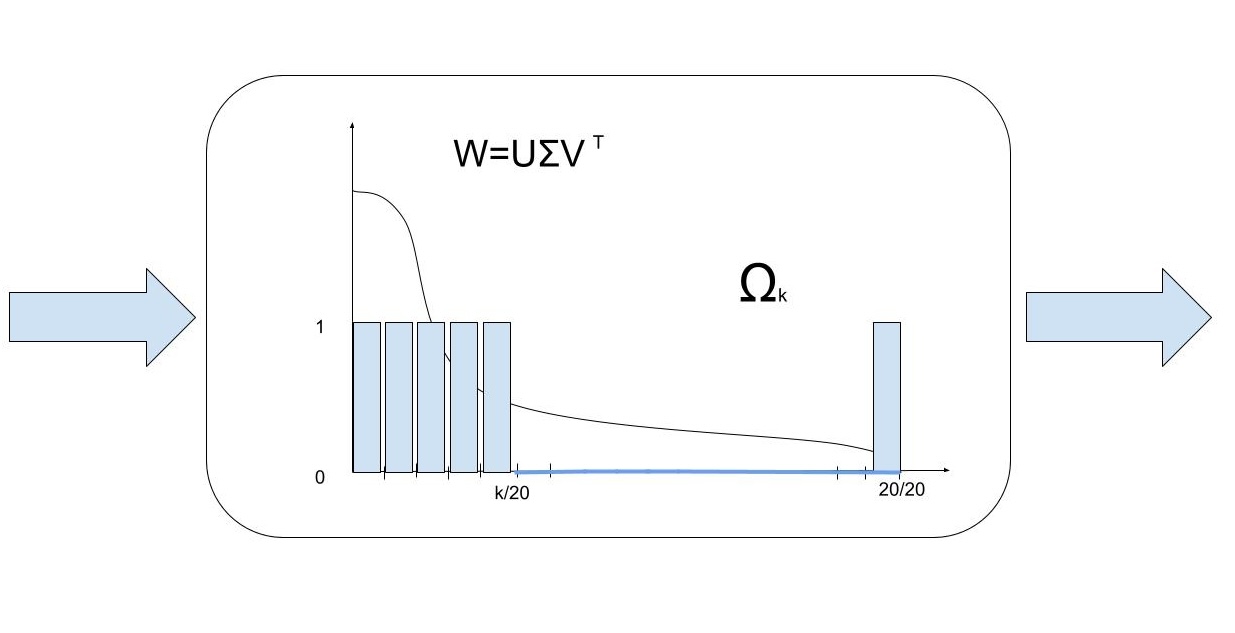}
    \caption{$\Omega$ filters project vectors on the head of the spectrum, but also on the tail end to preserve attention sinking.}
    \label{fig:omega_filter}
\end{figure}
The finding that dark signals are essential to enable attention sinks leads to the following question: what would be the impact of spectral filters that preserve dark signals but filter away one or more bands before them? Let $\Omega_{u,k}$ be a new family of spectral filters (Fig. \ref{fig:omega_filter}):
\begin{equation}
    \Omega_{u,k} = ([V_{u,1:k};V_{u,20:20}])([V_{u,1:k};V_{u,20:20}])^{\top}
\end{equation}

\begin{figure}
    \centering
    \includegraphics[width=\linewidth]{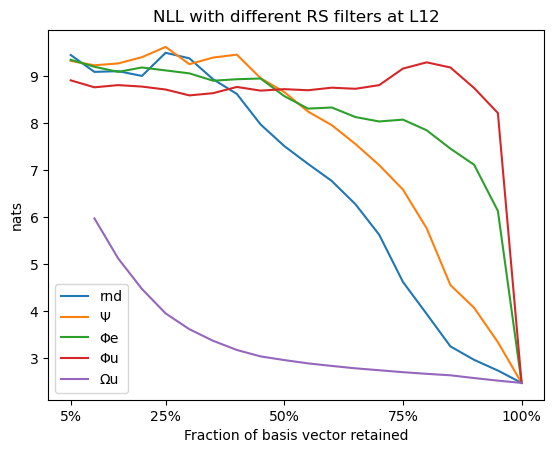}
    \caption{NLL by number of retained dimensions when applying different spectral filters at L12.}
    \label{fig:nll_13b_L12}
\end{figure}

Fig. \ref{fig:filter_13b_patch_bos_omega} (right) shows that low values of NLL are achieved even when masking a significant number of components. As an example (Fig. \ref{fig:nll_13b_L12}) NLL grows only from 2.47 to 2.74 when suppressing 25\% of the SVs by applying $\Omega_{u,14}$ after L12. We validated these results by repeating the experiment with a sample of code from the DeepMind Code Contest dataset \citep{doi:10.1126/science.abq1158} consisting of 117 prompts (14,641 tokens), Fig. \ref{fig:dmcc_filter_13b_rs_one}, and with LLaMa2 7B and 70B (Fig. \ref{fig:filter_7b_70b_rs_one}).

\subsection{The effect of inhibiting attention sinks on generation quality}
\label{sec:effects_on_generation_quality}

So far we have looked at the effect of spectral filters on the negative log-likelihood of prefixes that were fed as prompts to our models. It is interesting to also observe the impact on generated continuations when applying these same filters. 

\begin{table}[]
    \centering
    \renewcommand{\arraystretch}{1.5}
    \small
    \begin{tabular}{|p{6.8cm}|}\hline
         {\par \textbf{Prompt:} If you’re interested in Grizzly Bear viewing in and around the northern Gulf of Alaska, consider a chartered flight with Trygg Air Alaska based out of King Salmon with service to and from Anchorage.} \\\hline
         $\mathbf{\Psi (90\%)}$:  The EA, a GS, the E, A, for a total, the GS, the E, the S, the E, the S, the 15, the 16, the 15, the 16, the 15, the 17, the \\\hline
         \textbf{Rnd (90\%)}: What is the most beautiful thing that has ever been? What is the most beautiful thing that has ever been? It is not about the most beautiful thing that has ever been. The most beautiful thing that has ever been. It is not about the most beautiful thing that has ever been. It is not\\\hline
         $\mathbf{\Omega_{u}}$ (90\%): The Grizzly Bear is one of the largest land carnivores in the world. Males can reach a weight of 800 pounds and females can reach a weight of 400 pounds. The Grizzly Bear is one of the most powerful predators in North America  \\\hline
    \end{tabular}
    \caption{Sample generations by LLaMa2 13B with spectral filters applied right after L3.}
    \label{tab:generation_examples}
\end{table}

Layers 2-4 are the most fragile to random and $\Psi$ filtering (Tab. \ref{tab:generation_examples}), while generations remain coherent with $\Omega$ filtering irrespective of where the filter is applied, when 10-20\% of the singular vectors are suppressed. We notice that the application of $\Psi$ filters often results in the model entering repetitive patterns. This is consistent with the possibility that attention heads largely copy representations from the RSs of previous tokens, and inhibiting attention sinking results in over-copying. More examples can be found in App. \ref{sec:generation_examples}.

\section{Dark Signals and Attention Bars}
\label{sec:dark_signals_and_attention}

\begin{figure}
    \centering
        \includegraphics[width=0.7\linewidth]{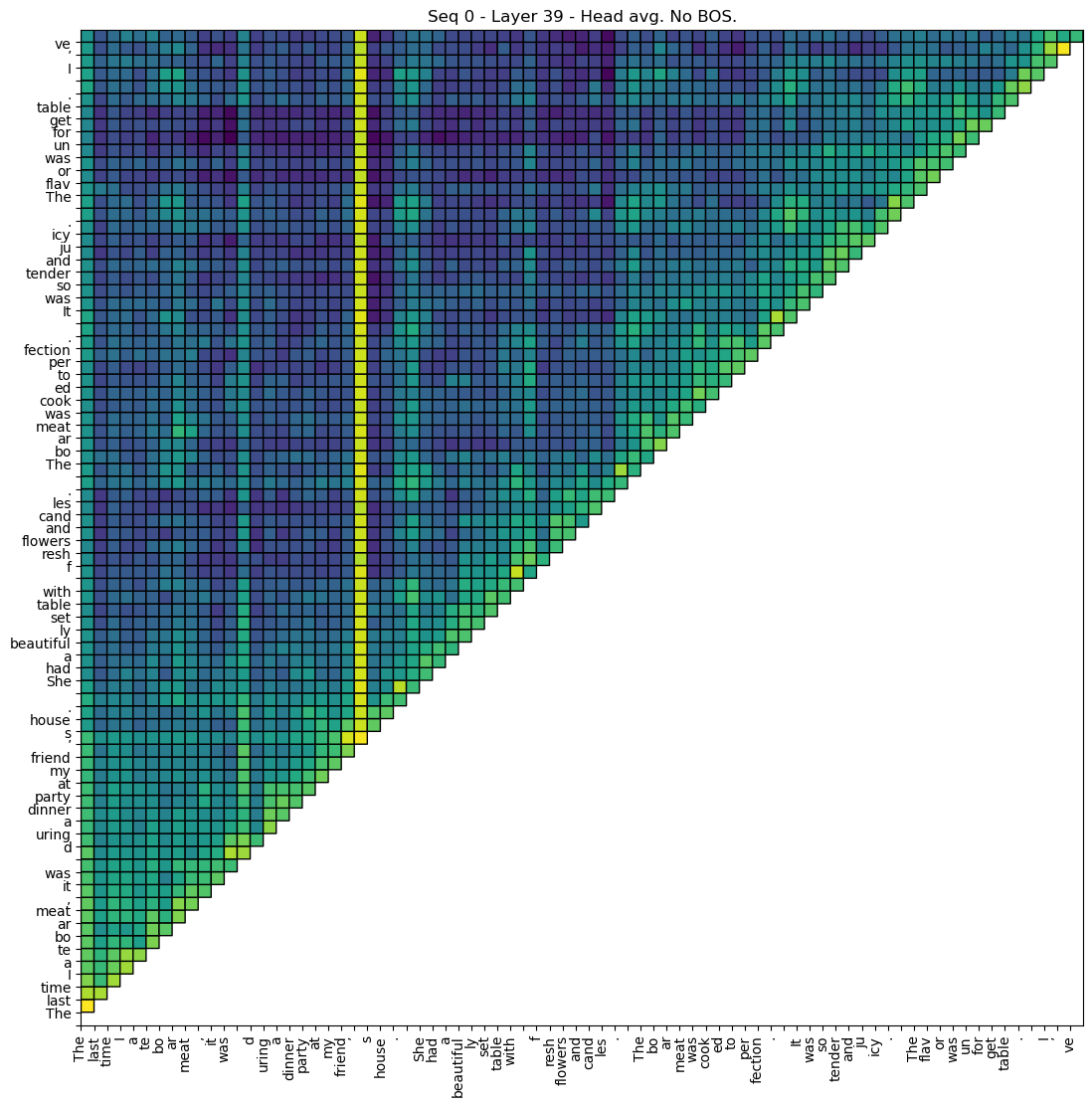}
        \label{fig:seq0-attn_L39}
    \caption{Attention matrix at L39 averaged across heads for a sample. Brighter columns correspond to tokens that are attended by many following tokens (LLaMa2 13B, Token 0 is removed for readability).}
    \label{fig:attn_head}
\end{figure}

Inspecting attention matrices in our sample we see that there are frequently a few tokens that also receive a large amount of attention, giving rise to characteristic \textit{attention bars} (e.g. Fig. \ref{fig:attn_head}).
We hypothesized that such bars could correspond to tokens whose residual streams are similar to that of Token 0, and therefore become additional attention sinks.

We define High-Mean Low-Variance (HMLV) token-layer pairs those with mean received attention over over subsequent tokens above $\tau_{\mu} = .018$ and variance below $\tau_{\sigma} = .01$. We ignore layers 0-3, since the attention sink is not in place yet, the BoS token, and the last 4 tokens of each sequence. This heuristic yields 12,236 HMLV token-layer pairs for LLaMa2 13B on ccnet-405 out of a total of 404,748 token-layer pairs for the same tokens and layers considered. We selected threshold manually so that most HMLV tokens appear as ``attention bars'' in attention matrices.
\begin{figure}[t]
    \centering
    \includegraphics[width=\linewidth]{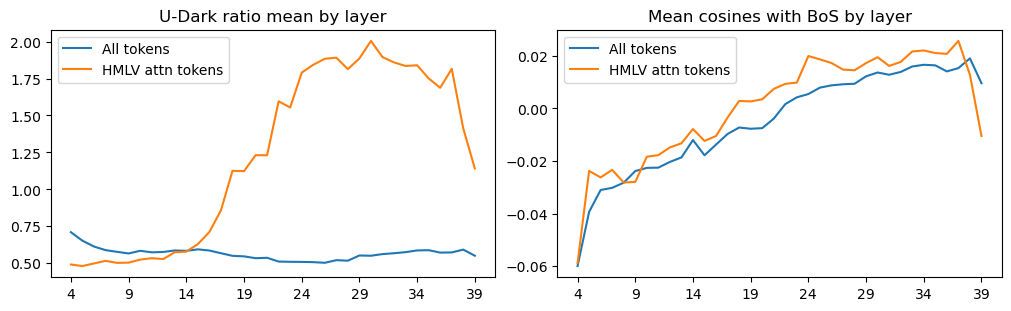}
    \caption{Mean U-Dark ratio (left) and cosine similarity with the BoS RS (right) for high-mean low-variance attention tokens and across all tokens, by layer.}
    \label{fig:hmlv_udark_cosines}
\end{figure}

We define the U-Dark ratio as a measure of the prevalence of U-Dark signals in a representation:

\begin{equation}
    udr(h_t^l) = \frac{||\Phi_{u,20:20}h_t^l||}{||(I -\Phi_{u,20:20})h_t^l||}
\end{equation}

\noindent
i.e. the norm ratio between the projection of the residual stream onto the U-dark space, and the projection on its orthogonal complement.

We measured average U-Dark ratio and cosine similarity with the BoS RS for HMLVs, and contrasted them with the same statistics taken over all token-layer pairs. Results are in Fig. \ref{fig:hmlv_udark_cosines}.

Starting around L13 the U-Dark ratio of HMLVs grows considerably, whereas it remains constantly below .75 for the overall population. At the same time, however, these tokens have cosine similarity with the BoS residual stream largely indistinguishable from the overall distribution, and close to zero. These results neither confirm nor invalidate the hypotheses that attention bars are auxiliary attention sinks. They seem to indicate that they occupy a different subspace \textit{within the dark subspace} from the BoS token, but they may still be attention sinks.

\section{Related work}
\label{sec:related_work}

In this work we adopt the framework introduced in \citep{elhage2021mathematical}, which highlights the central role played by the residual stream as working memory and communication channel across components, and hints at the notion of subspaces with differentiated functions. 
The \textit{logit lens} was first introduced by \citep{nostalgebraist2020}, and has been since used and extended in a number of interpretation studies, including \citep{dar2022analyzing,geva2022transformer,belrose2023eliciting,din2023jump,katz2023interpreting,voita2023neurons}. Of these, \citet{dar2022analyzing} analyse models projecting parameter matrices onto the vocabulary space, an approach we are also following in Section \ref{sec:parameter_properties}.

Our hypothesis that the subspace most orthogonal to the vocabulary representation could encode non-sparse latent features that should not interfere with the distribution over the immediate next token appeared supported by \citep{elhage2022toy}, where they show that, in basic models, important non-sparse features tend to be assigned principal components rather than being in superposition with sparse features. \citep{sharkey2022taking} showed that sparse autoencoders can recover a ground-truth assignment of features to linear subspaces by means of sparse autoencoders in an artificial and simplified setting and a 31M parameter LLM, with $d=256$; \citep{cunningham2023sparse} apply a similar method to larger Pythia 70M and Pythia 410M models. Our results with $\Omega$ filters with the much larger LLaMa2 models indicate that the residual stream could be used at less than full capacity, at least for some of the layers, raising the question of what model of superposition would best apply to them.

The concept of \textit{Attention Sinks} was introduced by \citet{xiao2023efficient}, who noticed that performance of streaming models dropped abruptly once the first tokens moved out of the attention sliding window context. We take here the next step in explaining how this process works, showing how it happens thanks to a very limited and specific portion of the spectrum.

\citet{sharma2023truth} show that it is possible to preserve and even improve the performance of the GPT-J \citep{gpt-j} model while dropping large portions of component matrices based on the SVD decomposition of the matrices themselves. We also focus on spectral decomposition, but using the spectrum of the unembedding matrix, in accordance with the logit lens approach and the intuition that it represents a common ground for all the components in a model.

\section{Conclusions and Future work}
\label{sec:conclusions}

A better understanding of the inner workings of transformer models is important to find strategies to make them safer. In this work we explored a novel way to look at transformers, extending the \textit{logit lens} approach into \textit{logit spectroscopy} by introducing \textit{spectral filters}. We explored the hypothesis that dark signals could be used to maintain global features while minimizing their interference with the next token, but we discovered that the main role of dark signals is as collectors for heads in need of an attention sink. We reconstructed how attention sinking works in LLaMa2 models, and we showed that, as long as attention sinking is preserved, they still achieve low loss even when significant portions of the unembedding spectrum are suppressed. Finally, we found a positive correlation between the attention received by a token and the relative prevalence of dark signals in its residual stream, especially in the upper layers.

The results on spectral filtering with dark signal preservation, combined with the observation that transformers are invariant to basis changes in the residual stream, suggest that it could be possible to first "canonicalize" a model so that its residual stream dimensions match the $V_u$ columns, and then compress away from components dimensions that are low in the spectrum but are not used for attention sinking. We consider such a form of \textit{spectral compression} an interesting direction for future work.

Finally, our exploration of the connection between ``attention bars'', prevalence of dark signals, and similarity with the Token 0 residual stream did not give a definite answer to the question of whether attention bars are additional attention sinks or have some other function. Increasing the granularity of the spectral bands in the dark subspace is a promising approach to solve this puzzle.

\section{Limitations}

We limited our attention to the LLaMa2 family of models, and only to the pretrained models prior to their instruction and safety fine-tuning. It is possible that other models do not realise attention sinks in the same way, or that they might not need them.\footnote{For example by using \textit{off by one} softmax \cite{miller2023offbyone}.}

The number of conditions tested in our detailed spectral filtering experiments meant that we could use only a text sample of limited size. While we did repeat some of the experiments using a code sample, we cannot exclude that there could be some differences when using samples from drastically different distributions, e.g. in languages using different scripts.

\section{Ethical Considerations}
\label{sec:ethical_considerations}

While our aim in studying the behaviour of LLMs is to make them more transparent, controllable, and trustworthy, it is conceivable that increased understanding could also benefit malicious agents intending to undertake harmful actions. Besides this generic considerations, we cannot see problematic ethical issues arising from this work.
\section*{Acknowledgements}

We thank the Transparent LLM team of Meta, in particular Lena Voita, Yihong Chen, Javier Ferrando, and Karen Hambardzumyan, for important suggestions and insightful discussions.

\bibliographystyle{assets/plainnat}
\bibliography{anthology,custom}

\begin{thebibliography}{29}
\providecommand{\natexlab}[1]{#1}
\providecommand{\url}[1]{\texttt{#1}}
\expandafter\ifx\csname urlstyle\endcsname\relax
  \providecommand{\doi}[1]{doi: #1}\else
  \providecommand{\doi}{doi: \begingroup \urlstyle{rm}\Url}\fi

\bibitem[Belrose et~al.(2023)Belrose, Furman, Smith, Halawi, Ostrovsky,
  McKinney, Biderman, and Steinhardt]{belrose2023eliciting}
Nora Belrose, Zach Furman, Logan Smith, Danny Halawi, Igor Ostrovsky, Lev
  McKinney, Stella Biderman, and Jacob Steinhardt.
\newblock Eliciting latent predictions from transformers with the tuned lens.
\newblock \emph{arXiv preprint arXiv:2303.08112}, 2023.

\bibitem[Cunningham et~al.(2023)Cunningham, Ewart, Riggs, Huben, and
  Sharkey]{cunningham2023sparse}
Hoagy Cunningham, Aidan Ewart, Logan Riggs, Robert Huben, and Lee Sharkey.
\newblock Sparse autoencoders find highly interpretable features in language
  models.
\newblock \emph{arXiv preprint arXiv:2309.08600}, 2023.

\bibitem[Dar et~al.(2022)Dar, Geva, Gupta, and Berant]{dar2022analyzing}
Guy Dar, Mor Geva, Ankit Gupta, and Jonathan Berant.
\newblock Analyzing transformers in embedding space.
\newblock \emph{arXiv preprint arXiv:2209.02535}, 2022.

\bibitem[Din et~al.(2023)Din, Karidi, Choshen, and Geva]{din2023jump}
Alexander~Yom Din, Taelin Karidi, Leshem Choshen, and Mor Geva.
\newblock Jump to conclusions: Short-cutting transformers with linear
  transformations.
\newblock \emph{arXiv preprint arXiv:2303.09435}, 2023.

\bibitem[Elhage et~al.(2021)Elhage, Nanda, Olsson, Henighan, Joseph, Mann,
  Askell, Bai, Chen, Conerly, DasSarma, Drain, Ganguli, Hatfield-Dodds,
  Hernandez, Jones, Kernion, Lovitt, Ndousse, Amodei, Brown, Clark, Kaplan,
  McCandlish, and Olah]{elhage2021mathematical}
Nelson Elhage, Neel Nanda, Catherine Olsson, Tom Henighan, Nicholas Joseph, Ben
  Mann, Amanda Askell, Yuntao Bai, Anna Chen, Tom Conerly, Nova DasSarma, Dawn
  Drain, Deep Ganguli, Zac Hatfield-Dodds, Danny Hernandez, Andy Jones, Jackson
  Kernion, Liane Lovitt, Kamal Ndousse, Dario Amodei, Tom Brown, Jack Clark,
  Jared Kaplan, Sam McCandlish, and Chris Olah.
\newblock A mathematical framework for transformer circuits.
\newblock \emph{Transformer Circuits Thread}, 2021.
\newblock \url{https://transformer-circuits.pub/2021/framework/index.html}.

\bibitem[Elhage et~al.(2022)Elhage, Hume, Olsson, Schiefer, Henighan, Kravec,
  Hatfield-Dodds, Lasenby, Drain, Chen, Grosse, McCandlish, Kaplan, Amodei,
  Wattenberg, and Olah]{elhage2022toy}
Nelson Elhage, Tristan Hume, Catherine Olsson, Nicholas Schiefer, Tom Henighan,
  Shauna Kravec, Zac Hatfield-Dodds, Robert Lasenby, Dawn Drain, Carol Chen,
  Roger Grosse, Sam McCandlish, Jared Kaplan, Dario Amodei, Martin Wattenberg,
  and Christopher Olah.
\newblock Toy models of superposition.
\newblock \emph{Transformer Circuits Thread}, 2022.
\newblock \url{https://transformer-circuits.pub/2022/toy_model/index.html}.

\bibitem[Geva et~al.(2020)Geva, Schuster, Berant, and
  Levy]{geva2020transformer}
Mor Geva, Roei Schuster, Jonathan Berant, and Omer Levy.
\newblock Transformer feed-forward layers are key-value memories.
\newblock \emph{arXiv preprint arXiv:2012.14913}, 2020.

\bibitem[Geva et~al.(2022)Geva, Caciularu, Wang, and
  Goldberg]{geva2022transformer}
Mor Geva, Avi Caciularu, Kevin~Ro Wang, and Yoav Goldberg.
\newblock Transformer feed-forward layers build predictions by promoting
  concepts in the vocabulary space.
\newblock \emph{arXiv preprint arXiv:2203.14680}, 2022.

\bibitem[Katz and Belinkov(2023)]{katz2023interpreting}
Shahar Katz and Yonatan Belinkov.
\newblock Interpreting transformer's attention dynamic memory and visualizing
  the semantic information flow of gpt.
\newblock \emph{arXiv preprint arXiv:2305.13417}, 2023.

\bibitem[Li et~al.(2022)Li, Choi, Chung, Kushman, Schrittwieser, Leblond,
  Eccles, Keeling, Gimeno, Lago, Hubert, Choy, de~Masson~d’Autume,
  Babuschkin, Chen, Huang, Welbl, Gowal, Cherepanov, Molloy, Mankowitz, Robson,
  Kohli, de~Freitas, Kavukcuoglu, and Vinyals]{doi:10.1126/science.abq1158}
Yujia Li, David Choi, Junyoung Chung, Nate Kushman, Julian Schrittwieser,
  R{\'e}mi Leblond, Tom Eccles, James Keeling, Felix Gimeno, Agustin~Dal Lago,
  Thomas Hubert, Peter Choy, Cyprien de~Masson~d’Autume, Igor Babuschkin,
  Xinyun Chen, Po-Sen Huang, Johannes Welbl, Sven Gowal, Alexey Cherepanov,
  James Molloy, Daniel~J. Mankowitz, Esme~Sutherland Robson, Pushmeet Kohli,
  Nando de~Freitas, Koray Kavukcuoglu, and Oriol Vinyals.
\newblock Competition-level code generation with alphacode.
\newblock \emph{Science}, 378\penalty0 (6624):\penalty0 1092--1097, 2022.
\newblock \doi{10.1126/science.abq1158}.
\newblock \url{https://www.science.org/doi/abs/10.1126/science.abq1158}.

\bibitem[Meng et~al.(2022)Meng, Bau, Andonian, and Belinkov]{meng2022locating}
Kevin Meng, David Bau, Alex Andonian, and Yonatan Belinkov.
\newblock Locating and editing factual associations in gpt.
\newblock \emph{Advances in Neural Information Processing Systems},
  35:\penalty0 17359--17372, 2022.

\bibitem[Miller()]{miller2023offbyone}
Evan Miller.
\newblock Attention is off by one.
\newblock \url{https://www.evanmiller.org/attention-is-off-by-one.html}.
\newblock https://www.evanmiller.org/attention-is-off-by-one.html.

\bibitem[Nostalgebraist()]{nostalgebraist2020}
Nostalgebraist.
\newblock interpreting gpt: the logit lens.
\newblock
  \url{https://www.alignmentforum.org/posts/AcKRB8wDpdaN6v6ru/interpreting-gpt-the-logit-lens}.

\bibitem[Olsson et~al.(2022)Olsson, Elhage, Nanda, Joseph, DasSarma, Henighan,
  Mann, Askell, Bai, Chen, Conerly, Drain, Ganguli, Hatfield-Dodds, Hernandez,
  Johnston, Jones, Kernion, Lovitt, Ndousse, Amodei, Brown, Clark, Kaplan,
  McCandlish, and Olah]{olsson2022context}
Catherine Olsson, Nelson Elhage, Neel Nanda, Nicholas Joseph, Nova DasSarma,
  Tom Henighan, Ben Mann, Amanda Askell, Yuntao Bai, Anna Chen, Tom Conerly,
  Dawn Drain, Deep Ganguli, Zac Hatfield-Dodds, Danny Hernandez, Scott
  Johnston, Andy Jones, Jackson Kernion, Liane Lovitt, Kamal Ndousse, Dario
  Amodei, Tom Brown, Jack Clark, Jared Kaplan, Sam McCandlish, and Chris Olah.
\newblock In-context learning and induction heads.
\newblock \emph{Transformer Circuits Thread}, 2022.
\newblock
  \url{https://transformer-circuits.pub/2022/in-context-learning-and-induction-heads/index.html}.

\bibitem[Pimentel et~al.(2020)Pimentel, Valvoda, Maudslay, Zmigrod, Williams,
  and Cotterell]{pimentel2020information}
Tiago Pimentel, Josef Valvoda, Rowan~H Maudslay, Ran Zmigrod, Adina Williams,
  and Ryan Cotterell.
\newblock Information-theoretic probing for linguistic structure.
\newblock In \emph{Proceedings of the 58th Annual Meeting of the Association
  for Computational Linguistics}, pages 4609--4622. Association for
  Computational Linguistics, 2020.

\bibitem[Sharkey et~al.(2022)Sharkey, Braun, and Millidge]{sharkey2022taking}
Lee Sharkey, Dan Braun, and Beren Millidge.
\newblock Taking features out of superposition with sparse autoencoders.
\newblock In \emph{AI Alignment Forum}, 2022.

\bibitem[Sharma et~al.(2023)Sharma, Ash, and Misra]{sharma2023truth}
Pratyusha Sharma, Jordan~T Ash, and Dipendra Misra.
\newblock The truth is in there: Improving reasoning in language models with
  layer-selective rank reduction.
\newblock \emph{arXiv preprint arXiv:2312.13558}, 2023.

\bibitem[Shazeer(2020)]{shazeer2020glu}
Noam Shazeer.
\newblock Glu variants improve transformer.
\newblock \emph{arXiv preprint arXiv:2002.05202}, 2020.

\bibitem[Su et~al.(2023)Su, Ahmed, Lu, Pan, Bo, and Liu]{su2023roformer}
Jianlin Su, Murtadha Ahmed, Yu~Lu, Shengfeng Pan, Wen Bo, and Yunfeng Liu.
\newblock Roformer: Enhanced transformer with rotary position embedding.
\newblock \emph{Neurocomputing}, page 127063, 2023.

\bibitem[Todd et~al.(2023)Todd, Li, Sharma, Mueller, Wallace, and
  Bau]{todd2023function}
Eric Todd, Millicent~L Li, Arnab~Sen Sharma, Aaron Mueller, Byron~C Wallace,
  and David Bau.
\newblock Function vectors in large language models.
\newblock \emph{arXiv preprint arXiv:2310.15213}, 2023.

\bibitem[Touvron et~al.(2023)Touvron, Martin, Stone, Albert, Almahairi, Babaei,
  Bashlykov, Batra, Bhargava, Bhosale, et~al.]{touvron2023llama}
Hugo Touvron, Louis Martin, Kevin Stone, Peter Albert, Amjad Almahairi, Yasmine
  Babaei, Nikolay Bashlykov, Soumya Batra, Prajjwal Bhargava, Shruti Bhosale,
  et~al.
\newblock Llama 2: Open foundation and fine-tuned chat models.
\newblock \emph{arXiv preprint arXiv:2307.09288}, 2023.

\bibitem[Vaswani et~al.(2017)Vaswani, Shazeer, Parmar, Uszkoreit, Jones, Gomez,
  Kaiser, and Polosukhin]{vaswani2017attention}
Ashish Vaswani, Noam Shazeer, Niki Parmar, Jakob Uszkoreit, Llion Jones,
  Aidan~N Gomez, {\L}ukasz Kaiser, and Illia Polosukhin.
\newblock Attention is all you need.
\newblock \emph{Advances in neural information processing systems}, 30, 2017.

\bibitem[Voita and Titov(2020)]{voita2020information}
Elena Voita and Ivan Titov.
\newblock Information-theoretic probing with minimum description length.
\newblock In \emph{Proceedings of the 2020 Conference on Empirical Methods in
  Natural Language Processing (EMNLP)}, pages 183--196, 2020.

\bibitem[Voita et~al.(2019)Voita, Talbot, Moiseev, Sennrich, and
  Titov]{voita2019analyzing}
Elena Voita, David Talbot, Fedor Moiseev, Rico Sennrich, and Ivan Titov.
\newblock Analyzing multi-head self-attention: Specialized heads do the heavy
  lifting, the rest can be pruned.
\newblock \emph{arXiv preprint arXiv:1905.09418}, 2019.

\bibitem[Voita et~al.(2023)Voita, Ferrando, and Nalmpantis]{voita2023neurons}
Elena Voita, Javier Ferrando, and Christoforos Nalmpantis.
\newblock Neurons in large language models: Dead, n-gram, positional.
\newblock \emph{arXiv preprint arXiv:2309.04827}, 2023.

\bibitem[Wang and Komatsuzaki(2021)]{gpt-j}
Ben Wang and Aran Komatsuzaki.
\newblock {GPT-J-6B: A 6 Billion Parameter Autoregressive Language Model}.
\newblock \url{https://github.com/kingoflolz/mesh-transformer-jax}, May 2021.

\bibitem[Wenzek et~al.(2019)Wenzek, Lachaux, Conneau, Chaudhary, Guzm{\'a}n,
  Joulin, and Grave]{wenzek2019ccnet}
Guillaume Wenzek, Marie-Anne Lachaux, Alexis Conneau, Vishrav Chaudhary,
  Francisco Guzm{\'a}n, Armand Joulin, and Edouard Grave.
\newblock Ccnet: Extracting high quality monolingual datasets from web crawl
  data.
\newblock \emph{arXiv preprint arXiv:1911.00359}, 2019.

\bibitem[Xiao et~al.(2023)Xiao, Tian, Chen, Han, and Lewis]{xiao2023efficient}
Guangxuan Xiao, Yuandong Tian, Beidi Chen, Song Han, and Mike Lewis.
\newblock Efficient streaming language models with attention sinks.
\newblock \emph{arXiv preprint arXiv:2309.17453}, 2023.

\bibitem[Zhang and Sennrich(2019)]{zhang2019root}
Biao Zhang and Rico Sennrich.
\newblock Root mean square layer normalization.
\newblock \emph{Advances in Neural Information Processing Systems}, 32, 2019.

\end{thebibliography}

\clearpage
\newpage
\beginappendix

\section{Notation}
\label{sec:notation}

Table \ref{tab:notation} summarizes the notation used throughout the paper.

\begin{table*}[]
    \centering
    \renewcommand{\arraystretch}{1.5}
    \begin{tabular}{|l|l|}\hline
        V & Vocabulary \\
        L<X> & Layer X \\
        H<Y> & Head Y \\
        $d$ & Model dimension \\
        $d_h$ & Head dimension \\
        $d_m$ & MLP hidden layer dimension \\
        $h_t^l\in \mathcal{R}^d$ & Intermediate repr. for Token $t$ at layer $l$ \\
        $W_z$, $z\in\{1,2,3\}$ & MLP matrices ($W_1, W_3\in\mathcal{R}^{d\times d_m})$, $W_2\in\mathcal{R}^{d_m\times d}$\\ %
        $W_z$, $z\in\{k,q,v,o\}$ & Attention matrices ($W_q, W_k, W_v\in \mathcal{R}^{d \times d_h}$, $W_o\in\mathcal{R}^{d_h\times d}$)\\ %
        $W_z$, $z\in\{e,u\}$ & Embeddings and unembeddings ($\in \mathcal{R}^{|V| \times d}$) \\ %
        $V_y, y\in\{e,u\}$ & right singular vectors of emb. and unemb. matrices\\\hline

    \end{tabular}
    \caption{Notation for referring to model components.}
    \label{tab:notation}
\end{table*}

\section{Implementation details}
\label{sec:implementation_details}

Spectral filtering experiments were run on servers with up to 8 NVIDIA A100 GPUs, each with 80GB of memory, and were implemented modifying the public Llama code from https://github.com/facebookresearch/llama. Inference used nucleus sampling with the default parameters (top-p = .9, T = .6).  Each heat-map took up to 4 hours to complete (for the 70B models). 

\section{Additional spectral filtering experiments}
\label{app:additional_spectral_filtering_plots}

\begin{figure}[b]
    \centering
    \includegraphics[width=\linewidth]{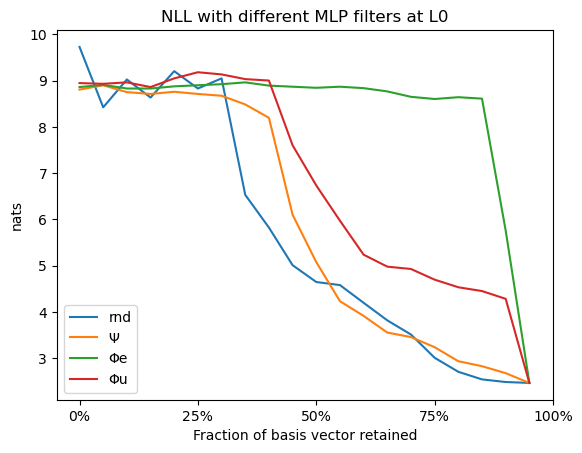}
    \caption{The effect of filtering 13B/L0/MLP with the filters defined in Section \ref{sec:spectral_filtering}.}
    \label{fig:mlp_filtering_L0}
\end{figure}

When filtering the output of the MLP of LLaMa2 13B at L0 with filters $\Phi_{e,1:k}$ log-likelihood remains poor (around 9) (Fig.\ref{fig:mlp_filtering_L0}) until about 90\% of $W_e$'s RSVs are retained. This indicates that the MLP in L0 exerts its effect by writing mostly into an E-dark subspace. The fact that the NLL recovers steadily starting from when about 40\% of the RSVs of $W_u$ are retained, and that the curve for the $\Psi_k$ filters is close to the random masking, show that this MLP writes mostly in the target unembedding space.

\begin{figure*}
    \centering
    \includegraphics[width=0.32\textwidth]{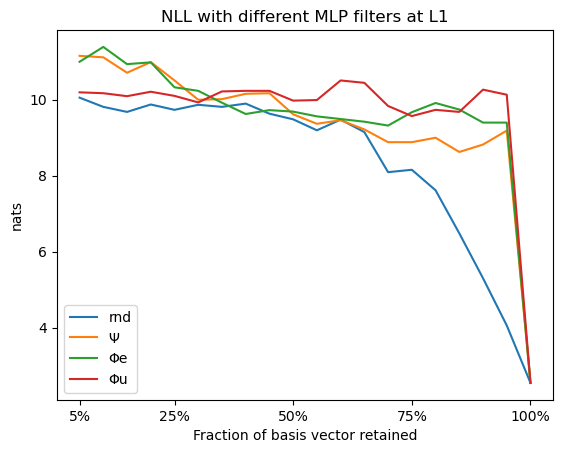}
    \includegraphics[width=0.32\textwidth]{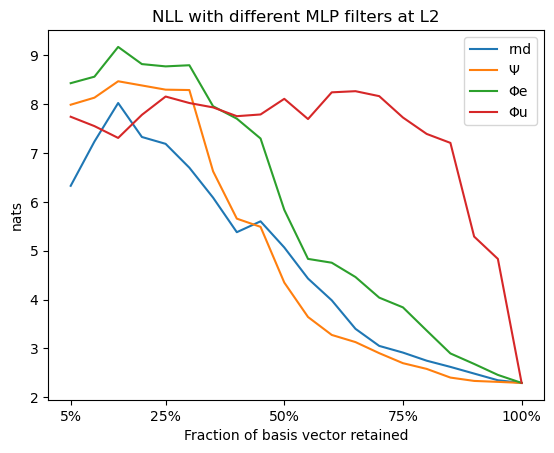}
    \includegraphics[width=0.32\textwidth]{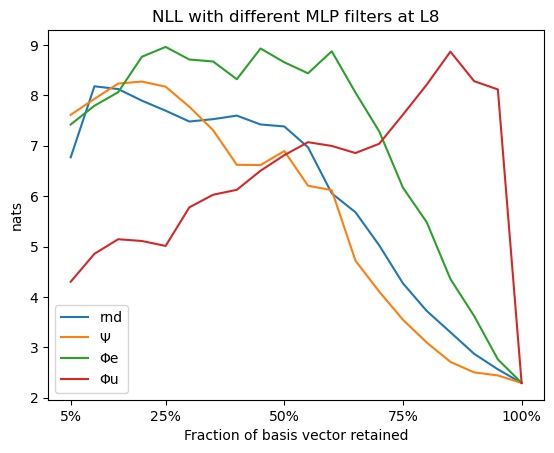}
    \caption{NLL filtering of MLPs in LLama2 7B (left) and 70B (middle) and (right) highlighting layers operating in the dark space.}
    \label{fig:filter_7b_70b_mlp_one}
\end{figure*}

\begin{figure*}
    \centering
    \includegraphics[width=0.32\textwidth]{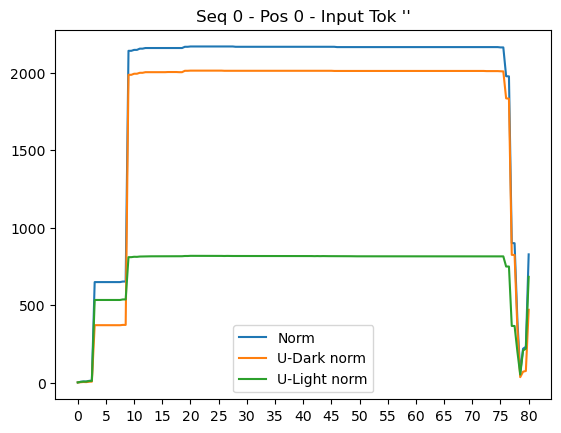}
    \includegraphics[width=0.32\textwidth]{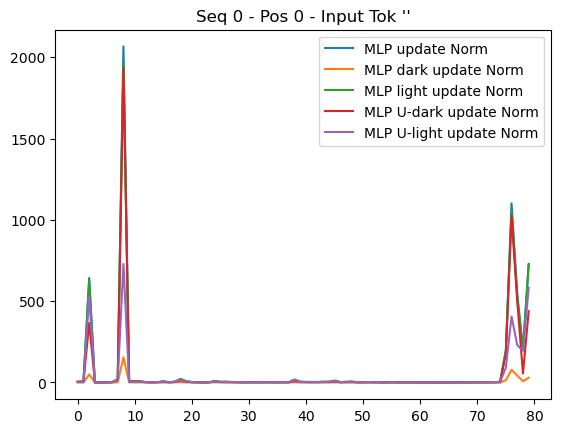}
    \includegraphics[width=0.32\textwidth]{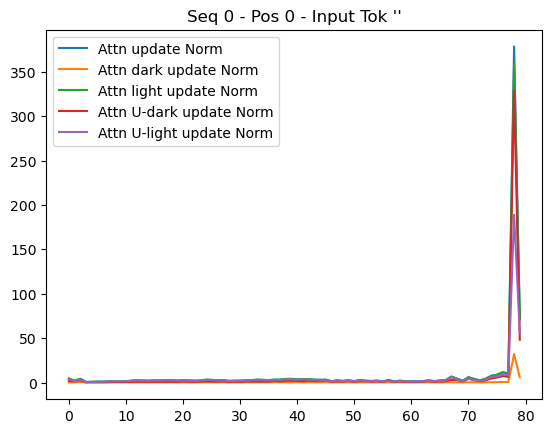}
    \caption{Token 0 residual stream in LLaMa2 70B. (left): RS norms by layer; (middle) norm of MLP contributions; (right) norm of MHA contributions.}
    \label{fig:70b_BoS}
\end{figure*}

Figure \ref{fig:filter_7b_70b_mlp_one} plots the NLL loss when applying spectral filters to the MLP at L1 in LLaMa2 7B, and at L2 and L8 in LLaMa2 70B. The loss profiles of component 7B/L1/MLP (left) are similar to those of 13B/L3/MLP: it exerts its influence in a subspace that is dark to both embeddings and unembeddings, and indeed it is the MLP that creates the attention sink vector in the Token 0 residual stream. 70B/L2/MLP and 70B/L8/MLP are the components responsible for the attention sink in 70b (Fig. \ref{fig:70b_BoS}). Unlike with 7B and 13B, the attention sink vector in Token 0 is U-Dark but not E-Dark, and is formed in two steps at L2 and L8.

\begin{figure*}
    \centering
    \includegraphics[width = 1.\textwidth]{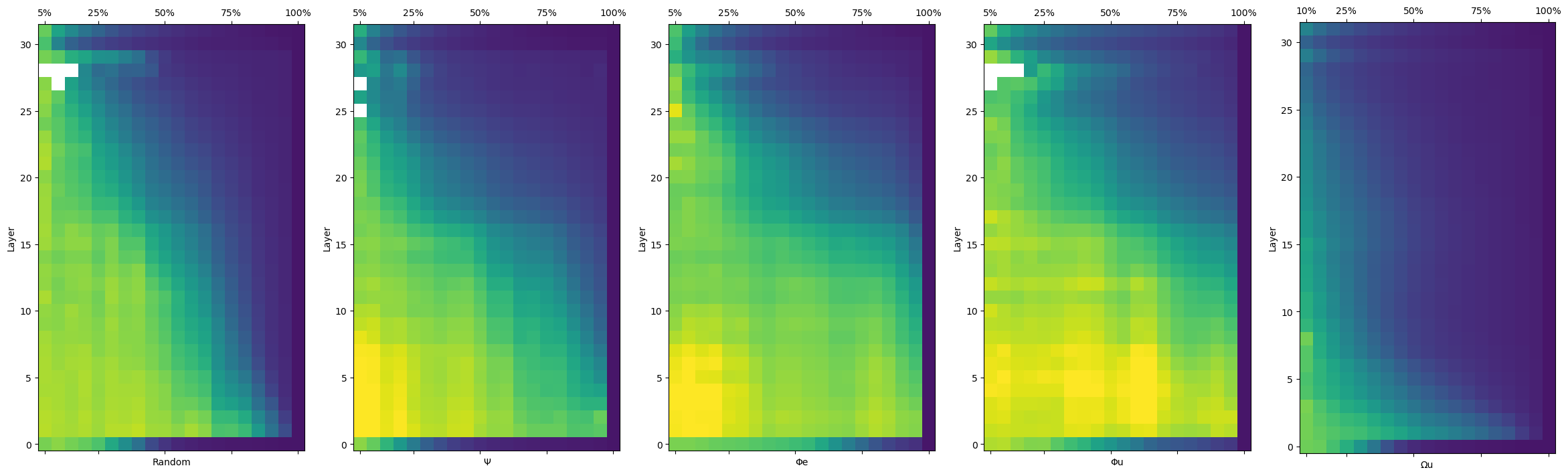}
    \includegraphics[width = 1.\textwidth]{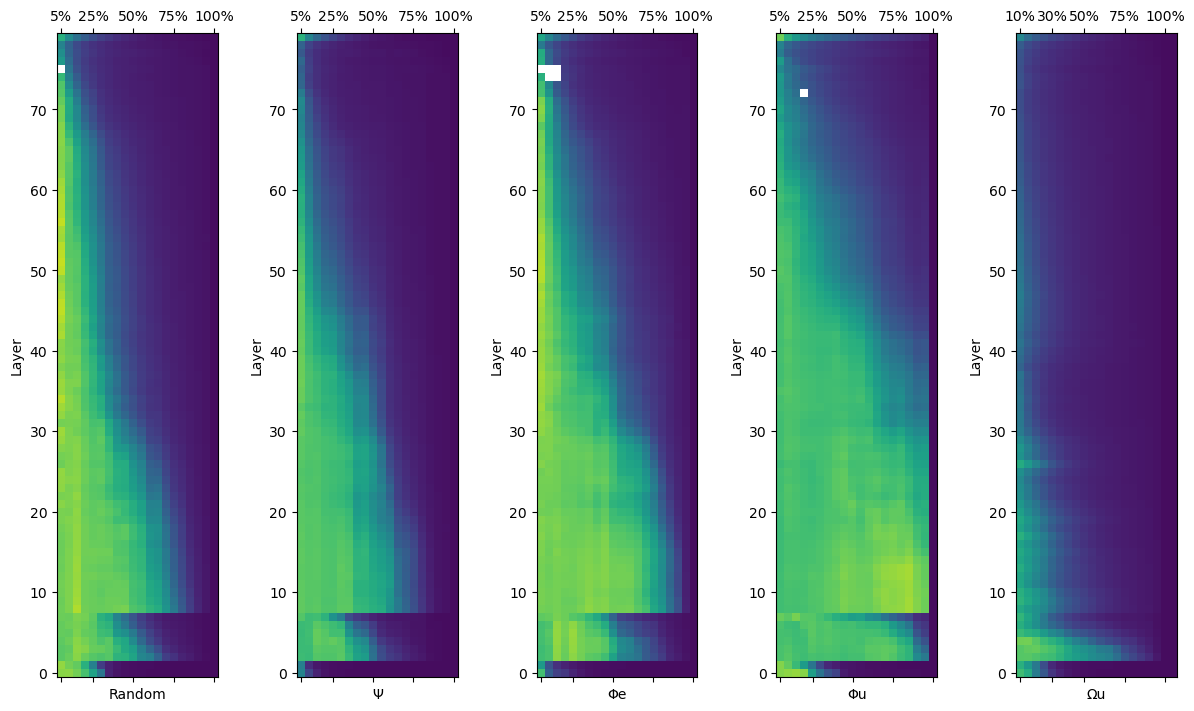}
    \caption{Negative Log-Likelihood when filtering the Residual Stream one layer at a time in LLaMa2 7B (top) and LLaMa2 70B (bottom). White cells correspond to configurations where NLL diverged for at least one of the samples.}
    \label{fig:filter_7b_70b_rs_one}
\end{figure*}

Figure \ref{fig:filter_7b_70b_rs_one} shows NLL loss heat-maps for LLaMa2 7B (top) and 70B (bottom) when spectral filters are applied to the RS at the layer on the Y axis.

\begin{figure*}
    \centering
    \includegraphics[width=\textwidth]{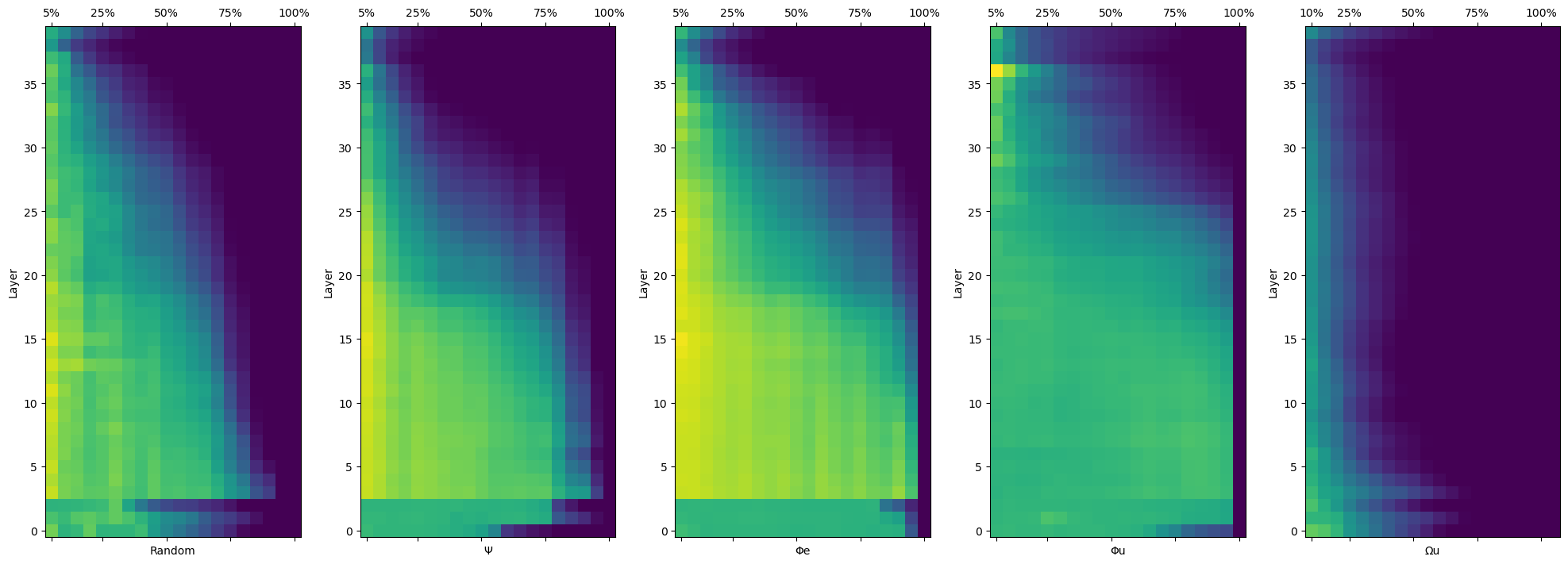}
    \caption{NLL filtering of MLPs in LLaMa2 13B on a sample of code from the DeepMind Code Contest dataset.}
    \label{fig:dmcc_filter_13b_rs_one}
    
\end{figure*}

Figure \ref{fig:dmcc_filter_13b_rs_one} displays similar heat-maps for LLaMa2 13B, using a sample of code fragments from the solutions of the DeepMind Code Contest dataset rather than ccnet-405.

\section{Generation examples}
\label{sec:generation_examples}

\begin{table*}[p]
    \footnotesize
    \centering
    \renewcommand{\arraystretch}{1.5}
    \begin{tabular}{|c|p{4.7cm}|p{4.7cm}|p{4.7cm}|}\hline
    \multicolumn{4}{|p{14cm}|}{\par \textbf{Prompt:} By Bart B. Van Bockstaele Mar 19, 2009 in Science} \\\hline
    & $\Psi$ (95\%) & Rnd (95\%) & $\Omega_{u}$ (95\%)\\\hline
        L0 & In this paper, we will discuss the differences between the currently most popularly used programming languages, C, C++, C\#, and Java. We will also discuss the advantages and disadvantages of each programming language. C is the oldest programming language of the four. C was developed by Dennis Ritchie in & A team of scientists from the University of Alberta, Canada, has discovered a new species of dinosaur. The new species is called Borealopelta markmitchelli. The scientists say that it is a close relative of the famous armored dinosaur Ankylosaurus. & In a recent study, researchers at the University of California, Riverside have found that the presence of an endocrine-disrupting chemical in the environment can alter the behavior of male fruit flies in a way that increases their chances of mating. The chemical in question is called meth\\\hline
        L2 & The most important thing is to find out how to get a good grip on the bart is the most important thing in the world. The bart is the most important thing in the world. the bart is the most important thing in the world. the bart is the most important thing in the world. & The giant planets Jupiter and Saturn are being observed by the Cassini spacecraft in the solar system. The images are showing the planets and their moons in a way that has never been seen before. The Cassini spacecraft is currently orbiting the planet Saturn, which it has been doing & A team of scientists from the University of California, San Diego, have discovered that a protein in the human body that is known to play a role in the progression of Parkinson's disease may also be a factor in the development of autism. The researchers have found that mutations in the protein\\\hline
        L3 & Van Bostis, 1969, 1969, 1969, 1969, 1969, 1969, 1969, 1969, 1969, 196 & By Bart B. Van Bockstaele Mar 19, 2009 in Science By Bart B. Van Bockstaele Mar 19, 2009 By Bart B. Van Bockstaele In the context of the current discussions on the future of the EU and the European Union, I would like to draw your attention to the following: & A new study by researchers at the University of California at San Francisco suggests that the use of acupuncture may actually reduce the risk of miscarriage. The study was based on data from 1,417 women who were pregnant and were treated at the San Francisco General Hospital between\\\hline
        L12 & The chemicals that make the brain think and the body work are the same ones that make the world go around, and the same ones that make the world go wrong. The universe is the great unknown, and it is a great unknown because it is the great unknown. The universe is a big place, and & A new study has found that a certain species of fish is able to survive in the most polluted waters of the world. The study, conducted by researchers from the University of Guelph, found that the fathead minnow, which is a small fish found in the Great Lakes, can surv & In a recent press release, the University of Leicester announced that its researchers have discovered a new way to get to the heart of the matter, literally. The researchers at the University of Leicester have discovered a new way to get to the heart of the matter, literally. The researchers\\\hline
        L32 & The oldest known animal to have ever lived on Earth, a 530-million-year-old sea creature, has been discovered in the Canadian Arctic. The creature, a 10-centimeter (3.9-inch) long worm, has been named the Hallucigenia & The "green" and "sustainable" movement is gaining momentum in the U.S. and around the world. The latest development is the Sustainable Urbanism conference in San Francisco, California. The "green" and "sustainable" movement is gaining momentum in the & The European Space Agency (ESA) is sending the first European astronaut into space since 1998. The mission is scheduled for the end of April. The ESA is sending the first European astronaut into space since 1998. The mission is scheduled for the end of April.\\\hline
        L38 & The discovery of a new species of bird in the Philippines has been announced by scientists from the Field Museum in Chicago. The new bird is called the Visayan warty pig, or the Visayan bleeding-heart pigeon. It was found on the island of Dinagat in & The European Space Agency (ESA) has released the first images of the 1323-pound (600 kg) GOCE satellite in orbit. The satellite is designed to map Earth's gravitational field with unprecedented accuracy. GOCE, or Gravity field and steady & The "Higgs boson" is the "God particle" that is needed to make the Standard Model complete. It has been missing for decades, but now the search is on in earnest. The Standard Model is a theory of elementary particles and their interactions. The model is based on the idea that\\\hline
    \end{tabular}
    \caption{Generations from LLaMa2 13B with Residual Stream filtered at one layer, comparing removing 5\% random projections, $\Psi$ filtering, and $\Omega_{u}$ filtering. Note that generations can be a mixture of true (but most likely never written by the author in the prompt) and hallucinated content.}
    \label{tab:psi_vs_rnd_rs_one_P19a}
\end{table*}

\begin{table*}[p]
    \centering
    \renewcommand{\arraystretch}{1.5}
    \footnotesize
    \begin{tabular}{|c|p{4.7cm}|p{4.7cm}|p{4.7cm}|}\hline
    \multicolumn{4}{|p{14cm}|}{\par \textbf{Prompt:} If you’re interested in Grizzly Bear viewing in and around the northern Gulf of Alaska, consider a chartered flight with Trygg Air Alaska based out of King Salmon with service to and from Anchorage.} \\\hline
    & $\Psi$ (90\%) & Rnd (90\%) & $\Omega_{u}$ (90\%)\\\hline
         L0 & \par Trygg Air Alaska is a small air taxi company based in King Salmon, Alaska. The company provides chartered flights from King Salmon to Anchorage, Alaska, which is the nearest city to King Salmon with an airport. The company is owned by a man named Try & \par The Alaska Department of Fish and Game (ADFG) has been conducting an aerial survey of the Gulf of Alaska for Grizzly Bears since 2006. The survey is conducted each summer to estimate the number of bears in the Gulf. The survey area covers & \par The flight is a one-hour scenic trip that flies over the Alaska Peninsula and Bristol Bay. You’ll be able to view the Alaskan wilderness and Grizzly Bears in their natural habitat.
        The flight includes a stop at the Brooks Camp for a\\\hline
         L3 & \par The EA, a GS, the E, A, for a total, the GS, the E, the S, the E, the S, the 15, the 16, the 15, the 16, the 15, the 17, the & \par What is the most beautiful thing that has ever been? What is the most beautiful thing that has ever been? It is not about the most beautiful thing that has ever been. The most beautiful thing that has ever been. It is not about the most beautiful thing that has ever been. It is not & \par The Grizzly Bear is one of the largest land carnivores in the world. Males can reach a weight of 800 pounds and females can reach a weight of 400 pounds. The Grizzly Bear is one of the most powerful predators in North America \\\hline
         L4 & \par The ACS, or, the A.S.E. The A.S.S.A.S. on the E.S.A. The A.S.E.E.A.S. on the E.S.A. The A.S.S.A & \par The Grizzly Bear is a wild creature and not a domesticated pet, so when you’re dealing with a wild creature, you have to be careful with your approach and how you do it. The Grizzly Bear is a wild creature and not a domesticated pet, so when you’ & \par Trygg Air Alaska is a family owned and operated air service based out of King Salmon, Alaska. We offer flightseeing tours and charter services to the Alaska Peninsula, Bristol Bay, and the Aleutian Islands. We have been flying in the region since 19 \\\hline
         L6 & \par The cost is about \$1,500 per person, but the trip is about a month long. The Gulf of Alaska is the only place in the world the bears are a problem. The trip is the same, but the bears are not. The bears are the same & \par Trygg Air Alaska is a charter service offering air transportation between Anchorage, Alaska and King Salmon, Alaska. With a fleet of two Piper Cherokee 6 aircraft, Trygg Air Alaska offers an economical alternative to commercial airlines and private aircraft. The & \par Trygg Air Alaska is a small air charter service based out of King Salmon, Alaska. They provide flight services to and from Anchorage and the remote communities of King Salmon, Iliamna, and Bristol Bay. Trygg Air Alaska offers a range of services, including scen \\\hline
         L10 & \par The Tail, the Beans, and the Salmon The Tail, the Beings, and the Salmon The Tail, the Beings, and the Salmon The Tail, the Beings, and the Salmon The Tail, the Beings, and the Salmon & Trygg Air Alaska is a family owned business, with the same values as our Alaskan ancestors: honesty, integrity and hard work. We’ve been in the air since 2010 and we’ve flown more than 500,000 miles over the beautiful & \par  Trygg Air Alaska is a family owned and operated airline based in King Salmon, Alaska. Trygg Air Alaska has been providing charter services to the Grizzly Bear Viewing lodges in and around the Gulf of Alaska since 1989. We have the\\\hline
        L22 & A Grizzly Bear viewing charter with Try The best time to see bears is from the end of the first week of the 2018/19 salmon run, through to the end of the 2019/2020 salmon run. The sal & Grizzly Bear viewing is an amazing experience. The bears are beautiful, majestic and awe-inspiring. The bears are also dangerous, and there are a lot of things to consider before you set out to view them. Trygg Air Alaska is dedicated to providing & Grizzly Bear viewing can be arranged in conjunction with fishing trips or as a separate trip. Fly-out trips can be arranged to view the Grizzly Bears at Brooks Falls, McNeil River, Katmai National Park, Lake Clark National Park,
\\\hline
        L32 & Trygg Air Alaska offers chartered flights for the purpose of viewing grizzly bears in and around the northern Gulf of Alaska. The northern gulf of Alaska is the only place in the world where grizzly bears are known to catch and eat the endanger & Grizzly Bear Viewing from the air The Gulf of Alaska is home to a large population of Grizzly Bears. There are approximately 1000 bears in the area, which is considered one of the largest Grizzly Bear populations in the world. In addition & Grizzly Bear Viewing, Fishing and More. Trygg Air Alaska is a charter air service based out of King Salmon, Alaska. Trygg Air Alaska has been operating for over 20 years. Trygg Air Alaska offers charters to the Katmai National\\\hline
    \end{tabular}
    \caption{Generations from LLaMa2 13B with Residual Stream filtered at one layer, comparing removing 10\% random projections, $\Psi$ filtering, and $\Omega_{u}$ filtering.}
    \label{tab:psi_vs_rnd_rs_one_P18}
\end{table*}

\begin{table*}[p]
    \footnotesize
    \centering
    \renewcommand{\arraystretch}{1.5}
    \begin{tabular}{|c|p{4.7cm}|p{4.7cm}|p{4.7cm}|}\hline
    \multicolumn{4}{|p{14cm}|}{\par \textbf{Prompt:} Tuatara are lizard-like reptiles that can be traced back to the time of the dinosaurs. For the first time in 200 years, researchers have discovered a baby tuatara on mainland New Zealand.} \\\hline
    & $\Psi$ (95\%) & Rnd (95\%) & $\Omega_{u}$ (95\%)\\\hline
        L1 & The baby tuatara is a tuatara that is a tuatara that is a tuatara that is a tuatara that is a tuatara that is a tuatara that is a tuatara that is a tuatara that is a tuatara that is a tuatara & The tuatara is a reptile that is unique to New Zealand and has survived from the time of the dinosaurs. It is the only surviving member of the order Rhynchocephalia. The tuatara was once found throughout the North Island and offshore islands, but & The discovery was made by a team of researchers from the University of Auckland and the Department of Conservation. The team was conducting a survey of tuatara populations on the island of Raoul, when they came across a juvenile tuatara. The tuatara is a species of\\\hline
        L3 & The full, 200 year, 1,166,1,1661,1,1,1661, 1,1,1661,1,1,1661,1, 1,1661,1,1,1 & Tuatara are lizard-like reptiles that can be traced back to the time of the dinosaurs. For the first time in 200 years, researchers have discovered a baby tuatara on mainland New Zealand. This is the first time since the 1800  & Tuatara are a native species to New Zealand, but they are notoriously difficult to breed in captivity. Tuatara are lizard-like reptiles that can be traced back to the time of the dinosaurs. For the first time in 200 years, \\\hline
        L4 & The adult is the first time a baby has been found in the world. The first time a baby has been found in the world. The first time a baby has been found in the world. The first time a baby has been found in the world. The first time a baby has been found in the & Tuatara are lizard-like reptiles that can be traced back to the time of the dinosaurs. For the first time in 200 years, researchers have discovered a baby tuatara on mainland New Zealand. The tuatara, which has not yet been & Tuatara are found only in New Zealand, and are the only surviving members of the order Rhynchocephalia, which dates back to the time of the dinosaurs. They can be found on the North, South and Stewart Islands, but the tuatara found on the mainland are\\\hline
        L7 & A 2016, the world’s population reached the 7.5 billion mark, the same number that now lives on the world, and the same number that now lives on the world, and the same number that now lives on the world, and the same number that now lives on the world, & Tuatara are lizard-like reptiles that can be traced back to the time of the dinosaurs. For the first time in 200 years, researchers have discovered a baby tuatara on mainland New Zealand. The tuatara was found in a rocky & Tuatara are lizard-like reptiles that can be traced back to the time of the dinosaurs. For the first time in 200 years, researchers have discovered a baby tuatara on mainland New Zealand. The tuatara is a reptile that \\\hline
        L10 & The last time a live one was found, the world was a different place. The world was at the end of the 1800s, the age of the "1840s, the 1840s, the 1840s, the 184 & The tuatara, named Pine, was found on Stephens Island in the Cook Strait, 15 kilometres from the nearest mainland. The tuatara, named Pine, was found on Stephens Island in the Cook Strait, 15 kilometres from the nearest mainland. Ph & The baby tuatara was found on the shore of a lake in the Otago region of New Zealand. Tuatara are a unique species of reptile that is native to New Zealand. They are the only living members of their family, and are closely related to lizards. \\\hline
        L15 & The small, 5.5-inch-long, 4-year-old female was found by a group of children at the, "I, or the world, would have to be pretty stupid to think that there weren't more than one," he said. Tuatua is the only & The tuatara was discovered in 2017 at the Karori Wildlife Sanctuary in Wellington, New Zealand. It is the first of its kind to be found in the wild on mainland New Zealand since the 1800s. Tuatara are a species of l & The tuatara is the last remaining member of a reptile order that flourished around 200 million years ago. It has survived for so long because of its slow metabolism, which allows it to survive on a diet of insects, small birds, lizards, \\\hline
        L36 & The baby tuatara was discovered on a small island off the North Island. It is believed the tuatara was washed ashore by a storm. The tuatara is the only surviving member of a group of reptiles that lived during the time of the dinosaurs. The & The tuatara is a nocturnal reptile, and it is a lizard-like creature that can be traced back to the time of the dinosaurs. Tuatara are lizard-like reptiles that can be traced back to the time of the dinosaurs & The tuatara is a lizard-like reptile that is endemic to New Zealand. It is the only living member of the order Rhynchocephalia, which flourished in the Triassic and Jurassic periods, and it is the last surviving member of a lineage \\\hline
    \end{tabular}
    \caption{Generations from LLaMa2 13B with Residual Stream filtered at one layer, comparing removing 5\% random projections, $\Psi$ filtering, and $\Omega_{u}$ filtering. Note that generations can be a mixture of true and hallucinated content.}
    \label{tab:psi_vs_rnd_rs_one_P19b}
\end{table*}

\begin{table*}[p]
    \footnotesize
    \centering
    \renewcommand{\arraystretch}{1.5}
    \begin{tabular}{|c|p{4.7cm}|p{4.7cm}|p{4.7cm}|}\hline
    \multicolumn{4}{|p{14cm}|}{\par \textbf{Prompt:} Tuatara are lizard-like reptiles that can be traced back to the time of the dinosaurs. For the first time in 200 years, researchers have discovered a baby tuatara on mainland New Zealand.} \\\hline
    & $\Psi$ (80\%) & Rnd (80\%) & $\Omega_{u}$ (80\%)\\\hline
        L3 & When, “b/3,on,h,a The “t The A,e,c,t,a,w,d, The,a,d,th, The,th, The, The,t,d, The,d, on,& At the time of the early 2000-600, 2000 and 2000: 2000, 2000, 2000 and 2000: 2000, 2000, & Tuatara are one of the largest reptiles in the world and are native to New Zealand. However, due to predation, habitat loss and the introduction of mammalian predators, they are now limited to a few small islands. The tuatara is a small lizard-like re\\\hline
        L5 & The first “tone” is the one, the first, the one, the first, the S, the first, the, the, the, the, the, the, the, the, the, the, the, the, the, the, the, the, the, the, the, the & In the 1990s, the public is on the 100th and 100th. The 100th is the 100th. The 100th is the 100th. The 100th is the 1 & A baby tuatara has been found for the first time in 200 years, according to a new study. Tuatara are lizard-like reptiles that can be traced back to the time of the dinosaurs, according to the study. The last time a tuat \\\hline
        L7 & , in the, The , in the, the , in the, The , the , the, and , the , the, the, the, the, the, the, the, the, the, the, the, the, the, the, the, the, the, & Ali T and/ 2d. Ali T and 2d. Ali T and 2d. Ali T and 2d. A. Ali T and 2d. A. A. Ali T and 2d. A. A. & Tuatara are lizard-like reptiles that can be traced back to the time of the dinosaurs. For the first time in 200 years, researchers have discovered a baby tuatara on mainland New Zealand. Researchers from Massey University in New\\\hline
        L10 & a y/t, a right The half, but, the, the, the, the, the, the, the, the, the, the, the, the, the, the, the, the, the, the, the, the, the, the, the, the, the, the & A single baby tuatara is a rare sight. It’s been 18 years since a baby tuatou & The discovery was made by the Department of Conservation (DOC) and the University of Otago, which have been working on a project to monitor tuatara on the mainland. The last time a tuatara was found on the mainland was in the 1800s.\\\hline
        L14 & Tuatara are lizard-like reptiles that can be traced back to the time of the dinosaurs. For the first time in 200 years, researchers have discovered a baby tuatara on mainland New Zealand. The E on a the, on the, the, at, 200, with, from, on, the, the, the, the, the, the, the, the, the, the, the, the, the, the, the, the, the, the, the, the & The small, taura taura, is 3000 years old, and it is the most important reptile in the world. Taura Taura is a 3000-year-old reptile that is the most important re 3000 years old re & Tuatara are lizard-like reptiles that can be traced back to the time of the dinosaurs. For the first time in 200 years, researchers have discovered a baby tuatara on mainland New Zealand. The discovery of the tuatara, which is\\\hline
        L27 & The newborn, which is only the 13th of its kind, was discovered on the “danger” Steph’, an, an, an, an, and the, and the, and the, and the, and the, and the, and the, and the, and the, and the & New Zealand'-Tiawan The island nation of New Zealand is a long way from the rest of the world, but that doesn't mean it's alone. It's the world's most isolated nation, but it's also home to some of the world's most unique and interesting & This is the first time in 200 years that a baby tuatara has been found on the mainland of New Zealand. The discovery was made by a group of researchers led by the University of Otago. The tuatara was found on the east coast of the North Island, \\\hline
        L36 & The Tuatara is an iconic New Zealand native. The reptiles can live for more than 100 years and are found on the main islands of the country and a few smaller islands. The last time a tuatara was born on the mainland was in 1895, & The small lizard, weighing just 40 grams, was found in the South Island's Kahurangi National Park. The species is usually found on the Chatham Islands, which are a group of small islands east of the South Island. The last time a tuatara & The discovery was made in a cave system in the remote Waitomo district, 300 kilometres south of Auckland. The young tuatara is about 10cm long and is the first recorded sighting of a juvenile tuatara on the mainland. The cave system is\\\hline
    \end{tabular}
    \caption{Generations from LLaMa2 13B with Residual Stream filtered at one layer, comparing removing 20\% random projections, $\Psi$ filtering, and $\Omega_{u}$ filtering. Note that generations can be a mixture of true and hallucinated content.}
    \label{tab:psi_vs_rnd_rs_one_P15}
\end{table*}

\end{document}